\documentclass[12pt]{article}
\usepackage{tabularx}
\usepackage{cite}
\usepackage{amsmath,amssymb,amsfonts}
\usepackage{graphicx,color}
\usepackage{textcomp}
\usepackage{xcolor}
\usepackage{hyperref}
\usepackage{placeins}
\usepackage{afterpage}
\usepackage{dblfloatfix}
\usepackage{xr}
\usepackage{array}
\usepackage{pdflscape}
\usepackage{booktabs}
\usepackage{pdflscape}
\usepackage{adjustbox}   
\usepackage{makecell} 

\newcolumntype{Y}{>{\raggedright\arraybackslash}X}

\newcolumntype{A}{>{\hsize=0.60\hsize\raggedright\arraybackslash}X} 
\newcolumntype{B}{>{\hsize=0.90\hsize\raggedright\arraybackslash}X} 
\newcolumntype{C}{>{\hsize=1.70\hsize\raggedright\arraybackslash}X} 
\newcolumntype{D}{>{\hsize=2.20\hsize\raggedright\arraybackslash}X} 

\usepackage{multirow}
\usepackage{algpseudocode}

\usepackage{multirow}
\usepackage{algorithm}
\newcommand{\cmark}{\ding{51}}
\newcommand{\xmark}{\ding{55}}
\usepackage{pifont}   
\usepackage{adjustbox}
\usepackage{rotating}
\def\BibTeX{{\rm B\kern-.05em{\sc i\kern-.025em b}\kern-.08em
    T\kern-.1667em\lower.7ex\hbox{E}\kern-.125emX}}
\AtBeginDocument{\definecolor{tmlcncolor}{cmyk}{0.93,0.59,0.15,0.02}\definecolor{NavyBlue}{RGB}{0,86,125}}

\title{HATL: Hierarchical Adaptive-Transfer Learning Framework for Sign Language Machine Translation}
\author{
Nada Shahin$^{1,2}$ and Leila Ismail$^{1,2}$%
\thanks{Corresponding author: Leila Ismail (\texttt{leila@uaeu.ac.ae}).}
}
\date{}
\begin{document}

\maketitle

\noindent
$^{1}$ Intelligent Distributed Computing and Systems (INDUCE) Lab, Department of Computer Science and Software Engineering, College of Information Technology, United Arab Emirates University, Abu Dhabi 15551, United Arab Emirates\\
$^{2}$ Emirates Center for Mobility Research, United Arab Emirates University, Abu Dhabi 15551, United Arab Emirates

\begin{abstract}
Sign Language Machine Translation (SLMT) aims to bridge communication between Deaf and hearing individuals. However, its progress is constrained by scarce datasets, limited signer diversity, and large domain gaps between sign motion patterns and pretrained representations. Existing transfer learning approaches in SLMT are static and often lead to overfitting. These challenges call for the development of an adaptive framework that preserves pretrained structure while remaining robust across linguistic and signing variations. To fill this void, we propose a Hierarchical Adaptive Transfer Learning (HATL) framework, where pretrained layers are progressively and dynamically unfrozen based on training performance behavior. HATL combines dynamic unfreezing, layer-wise learning rate decay, and stability mechanisms to preserve generic representations while adapting to sign characteristics. We evaluate HATL on Sign2Text and Sign2Gloss2Text translation tasks using a pretrained ST-GCN++ backbone for feature extraction and the Transformer and an adaptive transformer (ADAT) for translation. To ensure robust multilingual generalization, we evaluate the proposed approach across three datasets: RWTH-PHOENIX-Weather-2014 (PHOENIX14T), Isharah, and MedASL. Experimental results show that HATL consistently outperforms traditional transfer learning approaches across tasks and models, with ADAT achieving BLEU-4 improvements of 15.0\% on PHOENIX14T and Isharah and 37.6\% on MedASL. \end{abstract}



\section{INTRODUCTION}

Sign languages are the primary communication method for Deaf and hard-of-hearing (DHH) communities worldwide. These languages follow syntactic and grammatical structures that differ from spoken languages \cite{hall2019deaf}. This creates a growing need for Sign Language Machine Translation (SLMT) systems that support seamless communication between DHH and hearing individuals. Such technologies are critical in settings where human interpreters are unavailable. However, the development of SLMT systems is constrained by the scarcity of high-quality datasets, which are often limited in sample size, signer diversity, and annotation quality \cite{chen2022simple, fu2024signer}. Therefore, artificial intelligence (AI) models trained on sign language data tend to overfit and generalize poorly, as discussed in \cite{shahin2024glot,shahin2026adat}, establishing SLMT as a low-resource visual domain \cite{holmes2023scarcity, camgoz2018neural}. To overcome these limitations, recent research has adopted transfer learning to improve generalization, mitigate overfitting, and reduce the need for resource-intensive data collection \cite{hosna2022transfer}. 

Transfer learning is categorized into three types: 1) inductive, where knowledge is transferred from one domain to a related one; 2) transductive, where the domain is identical but labeled data is available only in the pretrained domain; and 3) unsupervised, where models rely on shared representations without labeled data \cite{hosna2022transfer, pan2009survey}. Among these, inductive transfer learning is the most suitable for SLMT. This is because it reuses general motion representations that were pretrained in related visual domains such as action recognition \cite{holmes2023scarcity}.

Large-scale inductive transfer learning has substantially improved performance in computer vision \cite{liu2022yolov5, xiao2022transfer, xue2021transfer, soleimani2021cross}, language understanding \cite{sung2022vl}, and speech processing \cite{al2025novel, cooper2020zero}. However, these advances are constrained by the high computational cost, the risk of negative transfer when feature distributions diverge, and the difficulty in balancing model capacity with limited data. These challenges are particularly pronounced in SLMT, where data scarcity and domain gaps restrict effective model adaptation. Moreover, sign language exhibits fine-grained spatial and temporal patterns, such as handshapes and non-manual signals, that differ from those in typical video domains, such as action recognition. 

Pretrained motion models encode a hierarchical structure that captures generic motion cues, temporal dependencies, and semantic information. In low-resource settings, such as SLMT, directly adapting these models by fine-tuning all parameters simultaneously can disrupt this hierarchy, as pretrained features are overwritten before the model converges. In contrast, limiting the extent to which the parameters can adapt prevents the capture of the linguistic and motion properties of the sign language. To mitigate these issues, existing approaches rely on manually unfreezing a fixed subset of pretrained layers. This selection is dataset-dependent, as the optimal set of activated layers varies with dataset scale and domain characteristics. As a result, generalization to unseen signers, sentences, and motion variations across SLMT datasets is limited \cite{hosna2022transfer,pan2009survey}. In this paper, we fill this void by introducing a novel Hierarchical Adaptive Transfer Learning (HATL) framework to improve adaptation stability and generalization in SLMT. HATL increases trainable capacity to gradually align more pretrained layers with sign language dynamics. This design provides a dynamic transfer mechanism that addresses overfitting and enables reliable adaptation to sign language. 

We divide SLMT into two categories: 1) Sign2Text, where sign videos are translated directly into spoken language text, and 2) Sign2Gloss2Text, which relies on gloss annotations, i.e., written sign representations, during translation. We evaluate HATL performance within these categories based on VisioSLR stages, an end-to-end framework for sign language recognition \cite{ismail2025visioslr}. We use a pretrained Spatio-Temporal Graph Convolutional Network (ST-GCN++) \cite{duan2022pyskl} for feature extraction and two translation models: the standard Transformer \cite{vaswani2017attention} and ADAT, an adaptive transformer-based model designed to capture the temporal nature of sign language sequences \cite{shahin2026adat}. We compare these models against two transfer learning approaches: 1) classical fine-tuning, where only the translation model is trained, and 2) full fine-tuning, where all pretrained parameters are updated simultaneously. We conduct experiments on three sign language datasets: the German RWTH-PHOENIX-Weather-2014T (PHOENIX14T) \cite{camgoz2018neural}, the Arabic Isharah \cite{alyami2025isharah}, and the American MedASL \cite{shahin2026adat} to assess the robustness of HATL across diverse languages and data characteristics. Our results demonstrate that HATL consistently improves translation across all settings, highlighting its potential as a robust transfer learning approach for SLMT.

The main contributions of this work are as follows.

- We propose HATL, a novel hierarchical adaptive transfer learning framework to improve SLMT.

- We compare HATL with classical and full fine-tuning baselines across three datasets, the German PHOENIX14T, the Arabic Isharah, and the American MedASL.

- We evaluate HATL, using the Transformer and ADAT, and compare the performance with existing works that use transfer learning for SLMT. Our experiments show that HATL outperforms existing static approaches. 

- We perform experiments to demonstrate how progressive adaptation improves translation.

The rest of the paper is organized as follows: Section II reviews related work. Section III describes the proposed framework. Section IV provides the experimental setup, performance evaluation, and results analysis. Section V concludes the paper and discusses future research directions.

\section{Related Work}
Several works have explored transfer learning across various application domains. In this section, we divide the related work into two categories: transfer learning applications in different domains and in SLMT.

Table~\ref{tab:transfer_different_related_work} presents the works that applied transfer learning in various domains \cite{liu2022yolov5, xiao2022transfer, xue2021transfer, soleimani2021cross, sung2022vl, al2025novel, cooper2020zero}. In computer vision, \cite{liu2022yolov5} demonstrates that pretrained Convolutional Neural Networks (CNN) backbones retain general spatial features under domain shifts. However, their approach relies on full fine-tuning, which increases training cost and sensitivity to overfitting. \cite{xiao2022transfer} decomposes synthetic-to-real LiDAR transfer into appearance and sparsity components, enabling partial adaptation through a learned point-cloud translator. This method relies on explicit domain alignment assumptions that may not generalize beyond the controlled settings. \cite{xue2021transfer} proposes a Transformer-based model for facial expression recognition using multi-attention regularization to prevent overfitting. However, this model depends on full fine-tuning, which increases training cost and limits scalability. \cite{soleimani2021cross} develops an adversarial cross-subject transfer learning framework for human-activity recognition. The framework reduces inter-user variability, but at the cost of training instability and sensitivity to source and target domains selection.

\begin{landscape}
\begin{table}[h]
\caption{Summary of Related Works on Transfer Learning for Different Domains}
\label{tab:transfer_different_related_work}
\centering
\scriptsize
\setlength{\tabcolsep}{4pt}
\renewcommand{\arraystretch}{1.02}
\setlength{\emergencystretch}{2em}

\begin{adjustbox}{width=\linewidth,totalheight=0.95\textheight,keepaspectratio,center}
\resizebox{\textwidth}{!}{%
\begin{tabular}{@{}
p{1.2cm} p{2.0cm} p{2.0cm} p{2.0cm} p{2cm}
p{3.0cm} p{2.0cm} p{1.8cm} p{2.2cm} p{2.2cm}
@{}}
\toprule
\multirow{2}{*}{\textbf{Work}} &
\textbf{Application} &
\multicolumn{2}{c}{\textbf{Architecture}} &
\multirow{2}{*}{\textbf{Dataset}} &
\multirow{2}{*}{\textbf{Unfreezing Strategy}} &
\multirow{2}{*}{\textbf{Adaptivity}} &
\multirow{2}{*}{\textbf{Framework}} &
\multirow{2}{*}{\textbf{Contributions}} &
\multirow{2}{*}{\textbf{Limitations}} \\
\cmidrule(lr){3-4}
& & \textbf{Extraction} & \textbf{Model} & & & & & & \\
\midrule

\cite{liu2022yolov5} &
UAV crop monitoring &
\multicolumn{2}{c}{YOLOv5l} &
Tassel Dataset &
Classic &
Static &
\xmark &
Shows that more domain-matched pretraining dataset improves performance &
Small dataset could be limiting the model performance and causing overfitting
\\

\cite{xiao2022transfer} &
3D LiDAR scene semantic segmentation &
\multicolumn{2}{c}{MinkowskiNet} &
SemanticKITTI, SemanticPOSS &
Not reported &
Static &
\xmark &
Shows that input-level transfer learning (like data augmentation) can improve segmentation performance &
Poor generalization beyond the used synthetic-real pairings
\\

\cite{xue2021transfer} &
Facial expression recognition &
Vision Transformer &
CNNs &
RAF-DB, FERPlus, AffectNet &
Not reported &
Static & 
\xmark &
Adds Multi-head Self-Attention Dropping to Transformer to improve relational learning &
Neglected local discriminative cues, poor generalization
\\

\cite{soleimani2021cross} &
Human activity recognition &
\multicolumn{2}{c}{Subject Adaptor GAN} &
Subset of Opportunity Challenge Dataset &
Partial (one component is updated per step) &
Dynamic (iterative adversarial adaptation) &
\xmark &
Proposes SA-GAN for cross-subject transfer learning &
Dataset-dependent and risk of inconsistent reproducibility and negative transfer
\\

\cite{sung2022vl} &
Vision-language tasks & 
CLIP &
BARTbase & 
VQAv2, GQA, NLVR2, MSCOCO captioning, TVQA, How2QA, TVC, YC2C & 
Partial (adapters, normalization layers, visual projection layer) &
Static &
\xmark &
Shows that adapters with weight sharing can match full fine-tuning performance while updating fewer parameters &
Fixed adapters may under-perform full tuning
\\

\cite{al2025novel} &
Speech &
Mel Frequency Cepstral Coefficients & 
Random Forest &
SceneFake & 
None (feature resuse) &
Static &
\xmark &
Introduces a transfer learning feature engineering method &
Poor generalization to unseen manipulations
\\

\cite{cooper2020zero} &
Multi-speaker text-to-speech &
Speaker embedding encoders &
Tacotron-based multi-speaker model &
VoxCeleb1+2 &
None (feature reuse) & 
Static &
\xmark &
Zero-shot application for speaker generalization &
Performance gap for zero-shot adaptation to unseen speakers
\\

\bottomrule
\end{tabular}
}
\end{adjustbox}
\end{table}
\end{landscape}

In language understanding, \cite{sung2022vl} proposes a parameter-efficient transfer via adapter modules for vision–language models, maintaining pretrained knowledge while reducing training overhead. However, the fixed adapters restrict flexibility in capturing layer transferability. In speech processing, \cite{cooper2020zero} demonstrates zero-shot transfer in text-to-speech using pretrained speaker embeddings, allowing adaptation without fine-tuning. However, this method shows degraded speaker similarity for unseen speakers. \cite{al2025novel} adopts feature-level transfer by reusing mel-frequency cepstral coefficients and Random Forest–based representations. This method achieves accurate classification and low computational cost, but with limited  knowledge transfer. 

In summary, existing approaches reveal a clear trade-off between adaptation strength, computational cost, and stability. Full fine-tuning improves task alignment but risks overfitting, while parameter-efficient and feature-level transfer improve efficiency but overlook the contribution of different layers to transferable representations. In contrast to the domains discussed above, SLMT combines action recognition and language understanding, requiring joint modeling of spatio-temporal visual dynamics and linguistic structure. Consequently, we focus on analyzing transfer learning approaches specifically designed for SLMT.

Table~\ref{tab:transfer_sign_related_work} compares the works that applied transfer learning to SLMT. These works are divided into two categories: 1) Sign2Text \cite{camgoz2018neural,guo2018hierarchical,guo2019hierarchical,camgoz2020sign,chaudhary2022signnet,yin2021simulslt,jin2022mc,yin-read-2020-better,li2022sign,chen2022two,chen2022simple,zhang2023sltunet}, and 2) Sign2Gloss2Text \cite{camgoz2018neural,camgoz2020sign,zhou2021spatial,chen2022simple,kan2022sign,yin2021simulslt,yin-read-2020-better,chen2022two,li2022sign,zhang2023sltunet,said2025adaptive}. 

\newcolumntype{Y}{>{\raggedright\arraybackslash}X}

\begin{landscape}
\begin{table}[p]
\caption{Summary of Related Works on Transfer Learning for Sign Language Machine Translation}
\label{tab:transfer_sign_related_work}
\centering
\scriptsize
\setlength{\tabcolsep}{2pt}
\renewcommand{\arraystretch}{1.02}
\setlength{\emergencystretch}{2em}

\begin{adjustbox}{width=\linewidth,totalheight=0.95\textheight,keepaspectratio,center}

\begin{tabular}{@{\extracolsep{\fill}}%
p{0.9cm}  
p{1.2cm}  
p{1.2cm}  
p{2.1cm}  
p{2.4cm}  
p{3.0cm}  
p{2.0cm}  
p{2.0cm}  
p{1.6cm}  
p{3.2cm}  
p{3.2cm}  
}
\toprule

\multirow{2}{*}{\textbf{Work}} &
\multicolumn{2}{c}{\textbf{Application}} &
\multicolumn{2}{c}{\textbf{Architecture}} &
\multirow{2}{*}{\textbf{Dataset}} &
\multirow{2}{*}{\textbf{Unfreezing}} &
\multirow{2}{*}{\textbf{Adaptivity}} &
\multirow{2}{*}{\textbf{Framework}} &
\multirow{2}{*}{\textbf{Contributions}} &
\multirow{2}{*}{\textbf{Limitations}} \\
\cmidrule(lr){2-3}\cmidrule(lr){4-5}
& \textbf{S2G2T} & \textbf{S2T} & \textbf{Extraction} & \textbf{Model} & & & & & & \\
\midrule

\cite{camgoz2018neural} &
\cmark & \cmark &
Spatial CNN &
RNN &
PHOENIX14T &
Classic &
Static &
\xmark &
Stage-wise transfer from recognition to translation  &
No linguistic feedback to encoder \\

\cite{camgoz2020sign} &
\cmark & 
\cmark &
EfficientNet &
Transformer &
PHOENIX14T &
Full &
Static &
\xmark &
Shared representations across recognition \& translation  &
Task dominance; sensitive to domain shift \\

\cite{guo2018hierarchical} &
\xmark & \cmark &
C3D CNN &
Hierarchical LSTM &
CSL &
Not reported &
Static &
\xmark &
Temporal transfer of motion features  &
High complexity; weak long-range linguistic alignment \\

\cite{guo2019hierarchical} &
\xmark & \cmark &
C3D CNN &
Hierarchical LSTM &
CSL &
Not reported &
Static &
\xmark &
Adaptive clip summarization with multi-modal fusion &
Complex training; limited scalability \\

\cite{zhou2021spatial} &
\cmark & \xmark &
VGG-11 &
BLSTM encoder + Segmented-Attention LSTM decoder &
PHOENIX14/14T, CSL &
Not reported &
Static &
\xmark &
Recognition-pretrained multi-cue transfer &
Cue contribution not controlled; sensitivity to noisy cues \\

\cite{chaudhary2022signnet} &
\xmark & \cmark &
ST-encoder &
Transformer &
PHOENIX14T &
Not reported &
Static &
\xmark &
Bidirectional Sign2Text/Text2Sign regularization via shared representations &
Dependent on keypoint quality \\

\cite{yin2021simulslt} &
\cmark & \cmark &
EfficientNet &
Transformer &
PHOENIX14T &
Full &
Static &
\xmark &
Sequential visual–gloss–text adaptation driven by latency constraints &
Latency--quality trade-off; stability under domain shift not addressed \\

\cite{jin2022mc} &
\cmark & \cmark &
CNN &
Transformer &
PHOENIX14T, CSL &
Not reported &
Static &
\xmark &
Fast adaptation to unseen signers &
Requires labeled adaptation; high computational overhead \\

\cite{yin-read-2020-better} &
\cmark & \cmark &
BiLSTM &
Transformer decoder &
PHOENIX14T &
Full &
Static &
\xmark &
Improved alignment via STMC and stronger temporal modeling &
Spoken language priors may mismatch sign grammar \\

\cite{chen2022simple} &
\cmark & \cmark &
S3D CNN &
Transformer &
PHOENIX14T, KETI &
Full &
Static &
\xmark &
Lightweight multi-modal transfer baseline &
Limited high-level adaptation; sensitive to modality noise \\

\cite{chen2022two} &
\cmark & \cmark &
Two S3D CNNs (RGB \& skeleton) &
Attention-based encoder + LSTM &
PHOENIX14T, CSL &
Full &
Static &
\xmark &
Cross-modal fusion enabling transfer &
Noise sensitivity; full tuning may destabilize convergence \\

\cite{li2022sign} &
\cmark & \cmark &
ViT (RGB) + Adaptive 3D GCN (Skeleton)  &
Transformer &
PHOENIX14T, CSL-Daily &
Full &
Static &
\xmark &
Strong motion modeling via skeletal dynamics + RGB fusion &
High compute; keypoint dependency \\

\cite{zhang2023sltunet} &
\cmark & \cmark &
SMKD &
Dual Transformer encoders (visual \& text) + Transformer decoder &
PHOENIX14T, CSL-Daily, DGS &
Classic &
Static &
\xmark &
Data-efficient unified modeling &
Linguistic mismatch from spoken MT data \\

This Work (HATL) &
\cmark & \cmark &
ST-GCN ++ &
Transformer \& Adaptive Transformer &
PHOENIX14T, Isharah, MedASL &
Progressive hierarchical unfreezing &
Dynamic &
\cmark &
Precision-aware transfer under domain shift &
Requires monitoring criteria; adds training protocol complexity \\
\bottomrule
\end{tabular}

\end{adjustbox}
\end{table}
\end{landscape}

In Sign2Text, \cite{camgoz2018neural} initializes CNN-LSTM encoders with pretrained sign representations, achieving faster convergence but low accuracy due to weak linguistic training. Hierarchical models \cite{guo2018hierarchical,guo2019hierarchical} use pretrained encoders and multi-level temporal modeling, improving robustness to motion variation despite increasing architecture complexity. \cite{camgoz2020sign} uses pretrained visual backbones with Connectionist Temporal Classification (CTC) supervision to stabilize training and improve fluency. However, it applies full fine-tuning which makes performance sensitive to domain shift. \cite{chaudhary2022signnet} shows that transferring keypoint-based features improves robustness across signers and domains. However, its performance is dependent on pose estimation quality. \cite{li2022sign} uses pretrained spatio-temporal graph neural networks to model skeletal dynamics, achieving strong motion modeling at the cost of higher computational complexity. \cite{yin2021simulslt} transfers pretrained action-recognition backbones for simultaneous Sign2Text translation, prioritizing low latency over high translation accuracy. \cite{jin2022mc} frames Sign2Text as a meta-learning problem. It enables rapid adaptation to unseen signers while increasing computational overhead. \cite{yin-read-2020-better} and \cite{zhou2021spatial} transfer pretrained RGB and pose representations, showing that multi-cue transfer improves performance. However, it increases training complexity and sensitivity to noise. \cite{chen2022two,chen2022simple} benefit from multi-modal pretraining but face scalability challenges. Lastly, \cite{zhang2023sltunet} jointly transfers knowledge across Sign2Text and related tasks using shared encoders and external translation corpora. This approach improves data efficiency but relying on spoken language pretraining introduces linguistic mismatch, as sign language grammar is not directly aligned with spoken syntax.

In summary, existing Sign2Text transfer learning methods show that pretrained visual and multi-modal representations are essential when data is scarce. Nevertheless, most approaches rely on static fine-tuning or domain-mismatched priors. These limitations motivate adaptive transfer strategies that selectively reuse pretrained knowledge while preserving the temporal and linguistic structures.

In Sign2Gloss2Text, transfer learning operates across two stages, where representations learned for sign recognition are reused to support text generation. \cite{camgoz2018neural} pretrains the visual encoder on isolated sign recognition and reuses it for gloss-based translation. This allows visual–linguistic knowledge to flow, improving alignment between motion features and gloss symbols. However, the separation between stages prevents linguistic feedback from influencing the encoder, limiting adaptability once the recognition stage converges. \cite{camgoz2020sign} mitigates this by learning shared representations between Sign2Gloss and Gloss2Text, enabling bidirectional transfer between both tasks. This approach improves visual–text alignment and training stability. However, it lacks control over task influence, leading to dominance of one modality. \cite{chen2022simple} shows that freezing pretrained encoders while fine-tuning fusion and decoding layers preserves low-level representations, but restricts higher-level adaptation to domain-specific syntax. \cite{zhou2021spatial} and \cite{kan2022sign} demonstrate that the reuse of cross-stage features across RGB, pose, and motion streams improves translation quality. However, the relative contribution of each modality to generalization remains underexplored. \cite{yin2021simulslt} introduces a dual translation model bridging Sign2Gloss2Text and Sign2Text by sequentially adapting visual, gloss, and text representations within a unified framework. This approach presents a shift from two-stage training toward integrated transfer. However, it is driven by latency constraints and does not explicitly address stability under large domain shifts. \cite{yin-read-2020-better}, \cite{zhang2023sltunet}, and \cite{said2025adaptive} integrate multi-task models by jointly training sign language recognition and translation. These models demonstrate that recognition-pretrained features can be reused for gloss-based translation. Nevertheless, they rely on static fine-tuning, increasing sensitivity to overfitting and unstable convergence in low-resource settings. \cite{chen2022two} and \cite{li2022sign} pretrain recognition components and then transfer their outputs to pretrained text decoders. Similar to previous works, these models use static full fine-tuning, increasing the risk of unstable convergence.

In summary, Sign2Gloss2Text transfer learning benefits strongly from recognition-pretrained representations, but existing approaches rely on stage-wise transfer, unconstrained joint training, or static fine-tuning. These limitations highlight the need for an adaptive transfer approach that regulates knowledge flow across stages while preserving sign language structure.

To conclude, existing Sign2Gloss2Text models rely on sequential transfer through gloss supervision, while Sign2Text approaches emphasize end-to-end multi-modal adaptation. Although recent works integrate cross-modal learning, transfer learning is still static, limiting stability under large domain shifts and constraining visual–linguistic alignment. To address these issues, we propose HATL, a hierarchical adaptive transfer learning framework that gradually adapts visual backbone layers. To our knowledge, no prior SLMT work provides such a dynamic transfer framework that progressively aligns feature representations while maintaining pretrained knowledge.

\section{Proposed Hierarchical Adaptive Transfer Learning (HATL) Framework}

\begin{figure}
\centerline{\includegraphics[width=4in]{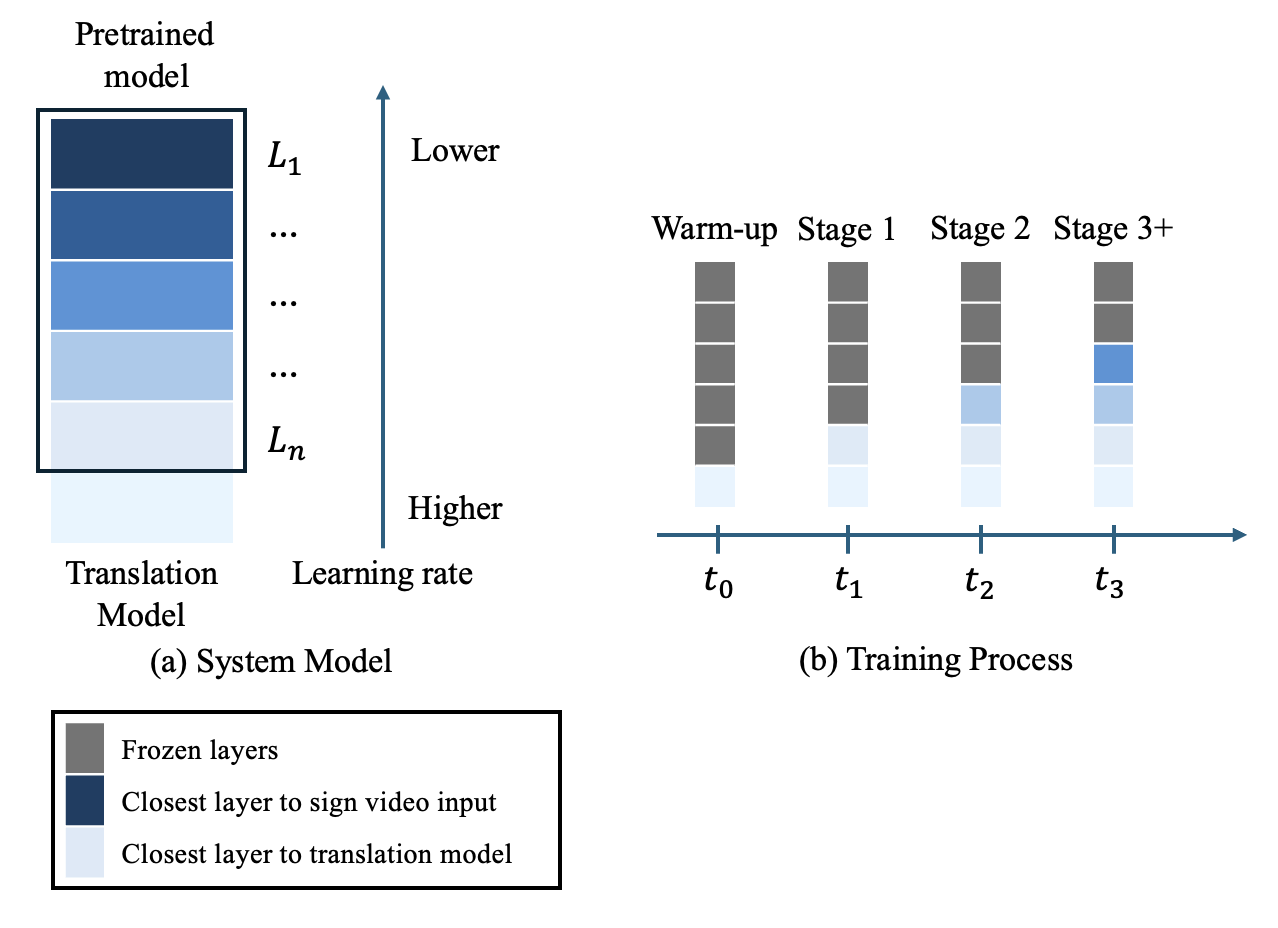}}
\caption{Overall HATL framework. \label{fig:hatl_overview}}
\end{figure}

In this section, we present HATL, a hierarchical performance-aware transfer learning framework for SLMT. HATL replaces static fine-tuning with dynamic hierarchical adaptation, progressively expanding trainable capacity based on performance behavior. It treats pretrained models as structured hierarchies, selectively and dynamically activating layers to adapt to sign language dynamics. We introduce the system model followed by the training process. Figure~\ref{fig:hatl_overview} illustrates the overall framework. Algorithm~\ref{alg:hatl} shows the HATL's algorithm.

\section{Algorithm}
\begin{algorithm}[!h]
\caption{HATL: Dynamic Adaptive Hierarchical Transfer Learning Framework}
\label{alg:hatl}
\begin{algorithmic}

\Require Training data $\mathcal{D}_{train}$, validation data $\mathcal{D}_{val}$;

backbone layers $L=\{L_1,L_2,\ldots,L_n\}$;
translation model $t$; thresholds $\Delta,\tau$;

size of moving–average window $k$; 
warmup period $\mathsf{warmup}$; patience $\mathsf{pat}$

\State Initialize trainable set $\mathcal{U}_0 \gets \{t\}$; freeze all $L_m$
\State Initialize optimizer with LLRD: 
      $LR_m = LR_t\cdot \alpha^{n-m}$
\State Initialize histories for $M(e)$, $\bar M(e)$, and $M'(e)$
\State $\text{pending\_release}\gets\emptyset$

\For{$e = 1$ to $E$}

    \If{$\text{pending\_release} \neq \emptyset$}
        \State Restore best-performing checkpoint
        \State Add $L_m$ to $\mathcal{U}_e$
        \State Rebuild optimizer with updated LLRD
        \State Apply cooldown; \quad $\text{pending\_release}\gets\emptyset$
    \EndIf

    \State Train $f(x;\Theta)$ on $\mathcal{D}_{train}$ using current $\mathcal{U}_e$
    \State Evaluate on $\mathcal{D}_{val}$ to compute $M(e)$
    \State Update $M'(e)$ and moving average $\bar M(e)$

    --- Release criterion for next backbone layer ---
    \If{$e > \mathsf{warmup}$}
        \If{$|M(e)-\bar M(e)| \le\Delta$\textbf{ and} $|M(e)-\bar M(e)| \le \tau$
        \textbf{ for } $\mathsf{pat}$ epochs}
            \State
            $\text{pending\_release}\gets L_{(m+1)}$
        \EndIf
    \EndIf

    --- Plateau-sensitive stopping rule ---
    \If{no improvement in $M(e)$ over several epochs}
        \State \textbf{break}
    \EndIf

    \State Gradually decay $\Delta \leftarrow 0.95\Delta$
\EndFor

\State \Return Best checkpoint $\Theta^\star$

\end{algorithmic}
\end{algorithm}

\subsection{System Model}
Figure~\ref{fig:hatl_overview}(a) presents the system model. We formulate SLMT using HATL as adapting a pretrained visual backbone and translation model to minimize the training objective while dynamically expanding the trainable parameters.

Given a sign language input sequence \(x\), the objective is to adapt a pretrained model \(f(\cdot;\Theta)\) to produce stable and effective predictions, as shown in Equation~\ref{eq:prediction}.
\[
\hat{y} = f(x;\Theta)
\tag{1}\label{eq:prediction}
\]
where \(\hat{y}\) is the SLMT model output.

We decompose the SLMT model into two main components: a pretrained backbone that serves as a visual feature extractor and a translation model that maps these features to linguistic outputs. Equation~\ref{eq:model} presents the full SLMT model.
\[
f(x;\Theta) = t\!\left(b(x;W_b);\,W_t\right)
\tag{2}\label{eq:model}
\]
where \(t(\cdot)\) is the translation model, \(W_t\) is its parameter, \(b(\cdot)\) is the pretrained backbone model, and \(W_b\) is its parameter.

 The backbone is further divided into an ordered hierarchy of layers, \(L=\left\{L_1,L_2,\ldots,L_n\right\}\), with \(L_1\) denoting the closest layer to the sign video input and \(L_n\) the closest layer to the translation model. At epoch \(e\), only a subset of parameters \(\mathcal{U}_e\subseteq\left\{L_1,L_2,\ldots,L_n,t\right\}\) is trainable. Training begins with \(\mathcal{U}_0=\left\{t\right\}\), such that only the translation model is trained while all backbone layers remain frozen. This decomposition provides the foundation for HATL’s progressive transfer mechanism.

\subsection{Training Process}
Figure~\ref{fig:hatl_overview}(b) shows the training process, which consists of: 1) Adaptive transfer learning, 2) Layer-wise learning rate control, 3) Stability mechanisms, and 4) Loss function. 
\subsubsection{Adaptive Transfer Learning}
HATL determines when to activate additional pretrained layers for adaptation through performance-aware adaptation. Let \(M(e)\) denote a validation metric in epoch \(e\) with \(M'(e)\) as the best value at epoch \(e\). To reduce per-epoch fluctuations, we compute a moving average over the previous epochs as shown in Equation~\ref{eq:moving_avg}.
\[
\bar{M}(e) = \frac{1}{k} \sum_{j=e-k+1}^{e} M(j)
\tag{3}\label{eq:moving_avg}
\]
where \(\bar{M}(e)\) is the smoothed validation metric at epoch \(e\), \(M(j)\) is the validation metric at epoch \(j\), \(j\) is the summation index over epochs, and \(k\) is the moving-average window size.

The decision to activate a backbone layer \(L_m\) is based on two criteria: (i) there is no significant improvement beyond a small margin \((\Delta)\) for several epochs, \(|M(e)\ -\bar{M}(e) |\le\Delta\), indicating convergence, and (ii) the improvement relative to the best observed performance is below a threshold \((\tau)\), \(|M(e)\ -\ \bar{M}(e)|\le\tau\). When both conditions are met, \(L_m\) becomes trainable at the start of the next epoch. This mechanism ensures that HATL expands trainable capacity only after the model has reached a stable optimization state. 

\subsubsection{Layer-wise Learning Rate Control}
Once a backbone layer is activated, HATL updates the optimizer to adjust the model parameters. In particular, it applies layer-wise learning rate decay (LLRD) \cite{howard2018universal} to assign different learning rates to different layers. For layer \(m\), the learning rate is set to:
\[
LR_m = LR_t \cdot \alpha^{n-m}
\tag{4}\label{eq:LLRD}
\]
where \(LR_t\) is the translation model learning rate and \(\alpha\in(0,1)\) is a decay factor.

This results in larger learning rates for layers closer to the translation model, leading to stronger adaptation, and smaller learning rates for layers closer to the sign video input, to preserve generic features. 

\subsubsection{Stability Mechanisms}
HATL incorporates several safeguards to ensure reliable progress during the adaptive transfer process. Training begins with a warmup phase, where only the translation model is being trained to ensure stability before transfer begins. Before activating any additional layer, HATL restores the model parameters from the best validation performance to prevent propagating unstable states. In addition, HATL applies a cooldown period after each layer activation, during which no further layers can be activated. Moreover, the threshold \(\Delta\) gradually decays, allowing the layer activation criterion to become more selective as learning slows. Finally, an early stopping rule terminates training when validation performance no longer improves.

\subsubsection{Loss Function}
HATL uses a weighted multi-objective loss function to supervise gloss alignment, text generation, and intermediate visual representations. Equation~\ref{eq:loss} defines the overall loss.
\[
\mathcal{L}={\omega_{CTC}\mathcal{L}}_{CTC}+{\omega_{CE}\mathcal{L}}_{CE}+\omega_{enc}\mathcal{L}_{enc}+\omega_{bb}\mathcal{L}_{bb}
\tag{5}\label{eq:loss}
\]
where \({L}_{CTC}\) is CTC loss for frame-to-gloss alignment, \({L}_{CE}\) is Cross-Entropy (CE) loss for training the autoregressive text decoder, and \({L}_{enc}\) and \({L}_{bb}\) apply frame-wise supervision in the encoder and backbone, respectively. The weights \(\omega_{CTC}\), \(\omega_{CE}\), \(\omega_{enc}\), and \(\omega_{bb}\)  control the influence of each component. For direct Sign2Text translation, the gloss alignment is disabled by setting \(\omega_{CTC} = 0\).

\paragraph{Connectionist Temporal Classification Loss} 
During Sign2Gloss2Text training, the encoder outputs frame-level gloss probabilities, as defined in Equation ~\ref{eq:gloss_pred}.
\[
p_g(c)=p(c \mid x)_g
\tag{6}\label{eq:gloss_pred}
\]
where \(p(c \mid x)_g\) is gloss predicted probability at frame \(x\).

Sign language often lacks explicit frame-level gloss boundaries. Therefore, we use CTC to model frame-to-gloss alignment. Given a target gloss sequence is \(Y=(y_1,\ldots,y_U)\), the CTC loss is defined in Equation~\ref{eq:ctc_loss}.
\[
\mathcal{L}_{CTC}
=
-\log\sum_{A\in B^{-1}(Y)}
P_{\mathrm{gloss-path}}(A\mid x)
\tag{7}\label{eq:ctc_loss}
\]
where \(B^{-1}(Y)\) denotes all valid alignment paths \(A=(A_1,\ldots,A_z)\) in the gloss sequence \(Y\). 

Equation~\ref{eq:gloss_align} defines the probability of gloss-alignment path.
\[
P_{\mathrm{gloss-path}}(A\mid x)=\prod_{g=1}^{G}p_g(A_g)
\tag{8}\label{eq:gloss_align}
\]
where \(p_g(A_g)\) is the predicted probability at frame \(g\).

\paragraph{Cross Entropy Loss} 
The autoregressive text decoder plays an increasingly larger role as more layers become trainable. Equation~\ref{eq:ce_loss} defines CE loss used for training text generation.

\begin{equation}
\label{eq:ce_loss}
\mathcal{L}_{\mathrm{CE}}
=
-\frac{1}{NS}
\sum_{i=1}^{N}\sum_{s=1}^{S}
\log p\!\left(
y_{i,s}^{\mathrm{text}}
\mid x_i, y_{i,<s}^{\mathrm{text}}
\right)
\tag{9}
\end{equation}

where \(N\) denotes the number of sign videos, \(S\) is the length of the target text sequence, \(i\) indexes the sign videos, \(s\) indexes token position within the target text sequence, and \(y^{\mathrm{text}}\) denotes the ground-truth target text token. The conditional distribution is parameterized by the translation model defined in Equation~\ref{eq:model}.

\paragraph{Encoder Frame-wise Supervision}
To maintain visual alignment within the encoder during unfreezing, HATL applies a frame-wise supervision at the encoder level. Equation~\ref{eq:enc_loss} presents this loss.

\begin{equation}
\label{eq:enc_loss}
\mathcal{L}_{\mathrm{enc}} = -\frac{1}{|\mathcal{M}|}\sum_{(i,s)\in\mathcal{M}}\log p\!\left(y_{i,s}^{\mathrm{enc}}\mid x_i\right)
\tag{10}
\end{equation}

where \(\mathcal{M}\) is the set of aligned frame indices used for supervision
and $y^{\mathrm{enc}}$ is the target encoder label.

\paragraph{Backbone Frame-wise Supervision}
HATL applies an additional loss directly to the backbone output to align the frames with the backbone. This component maintains stable feature transitions during progressive unfreezing. Equation~\ref{eq:bb_loss} defines this loss.
\begin{equation}
\label{eq:bb_loss}
\mathcal{L}_{\mathrm{bb}}
=
-\frac{1}{|\mathcal{M}|}
\sum_{(i,s)\in\mathcal{M}}
\log p\!\left(
y_{i,s}^{\mathrm{bb}}
\mid x_i
\right)
\tag{11}
\end{equation}

where $y^{\mathrm{bb}}$ is the target backbone.

These four loss components define HATL as a dynamic transfer learning framework. In particular, CTC enables early alignment while the backbone is mostly frozen. The encoder and backbone losses stabilize intermediate representations as layers are progressively unfrozen. The CE decoding dominates training once deeper layers become trainable. This design allows SLMT models to benefit from pretrained motion representations while gradually adapting to sign language translation.

 In summary, HATL adapts pretrained models to SLMT by dynamically expanding trainable layers based on validation performance. By combining hierarchical parameter access with performance-aware control, learning-rate regulation, and frame-wise supervision, HATL enables pretrained SLMT models to gradually specialize toward sign language translation while preserving robust visual representations learned during pretraining.

\section{Performance Evaluation}
\subsection{Experimental Environment}
\subsubsection{Datasets}
We evaluate the proposed HATL approach on three datasets: PHOENIX14T \cite{camgoz2018neural}, Isharah \cite{alyami2025isharah}, and MedASL \cite{shahin2026adat}. Table \ref{tab:datasets_char} summarizes the datasets characteristics. 

PHOENIX14T presents the highest linguistic variability due to its multi-signer nature, broad vocabulary, and long sequences. The large coefficients of variation (CV) in gloss and text indicate fluctuations in syntax and pacing, which challenge transfer stability and generalization. 

Isharah reflects a controlled setting. Despite including multiple signers, its sequences are short and highly consistent, with minimal CV in gloss and text lengths. This leads to evaluating primarily signer generalization rather than linguistic generalization.

MedASL's single-signer setting produces a uniform signing pattern and consistent sentence structure, while the medical domain introduces rich terminology and structured phrasing. This setting creates a controlled setting that maintains meaningful semantic variation. 

These datasets allow objective evaluation of HATL across multiple linguistic conditions. We use the predefined splits of  PHOENIX14T and Isharah, while we split MedASL into 80\% for 5-fold cross-validation and 20\% for testing.

\begin{table*}[h]
\caption{Dataset characteristics.}
\label{tab:datasets_char}
\centering
\setlength{\tabcolsep}{4pt}
\renewcommand{\arraystretch}{1.1}
\begin{adjustbox}{width=\linewidth,totalheight=0.95\textheight,keepaspectratio,center}

\begin{tabular}{llllllcccccc}
\toprule
\multicolumn{1}{c}{\multirow{2}{*}{Dataset}} &
\multicolumn{1}{c}{\multirow{2}{*}{Sign Language}} &
\multicolumn{1}{c}{\multirow{2}{*}{Domain}} &
\multicolumn{1}{c}{\multirow{2}{*}{\shortstack{Duration \\ (in hours)}}} &
\multicolumn{1}{c}{\multirow{2}{*}{\# Signers}} &
\multicolumn{1}{c}{\multirow{2}{*}{\# Videos}} &
\multicolumn{2}{c}{Vocabulary Size} &
\multicolumn{2}{c}{Avg. Length} &
\multicolumn{2}{c}{\shortstack{Coefficient \\ of Variation}} \\

\multicolumn{1}{c}{} &
\multicolumn{1}{c}{} &
\multicolumn{1}{c}{} &
\multicolumn{1}{c}{} &
\multicolumn{1}{c}{} &
\multicolumn{1}{c}{} &
Gloss &
Text &
Gloss &
Text &
Gloss &
Text \\
\midrule
PHOENIX14T & German & Weather & 10.53 & 9 & 8,257 & 1,115 & 3,000 & 7.66 & 13.77 & 0.49 & 1.06 \\
Isharah   & Arabic & Multiple & 10.14 & 15 & 7,500 & 388   & 758   & 4.19 & 3.88  & 0.06 & 0.04 \\
MedASL    & American & Medical & 5.56 & 1 & 2,000 & 1,142 & 1,682 & 5.35 & 8.60  & 0.12 & 0 \\
\bottomrule
\end{tabular}
\end{adjustbox}
\end{table*}

\subsubsection{Data Preprocessing}
We use a unified preprocessing setup to produce lightweight inputs. First, we resize input frames to $52{\times}65$ to ensure consistent resolution and efficient computation. Then, we extract keypoint representations from raw videos using MediaPipe \cite{google2024mediapipe}, capturing hand, face, iris, and upper-body landmarks. This is to remove unnecessary visual details and reduce input dimensionality while preserving essential motion and articulation cues needed for translation \cite{shahin2025towards}. We normalize and rescale the resulting coordinates. We then concatenate them across sentences and pad them for batch processing.

For the text output, we construct a vocabulary that includes start- and end-of-sentence tokens. We tokenize the sentences, map each token to its index, and pad the sequences to a fixed length to enable efficient batch-level training and evaluation.
\subsubsection{Model Development}
We evaluate HATL using a pretrained ST-GCN++ \cite{duan2022pyskl} for feature extraction and the Transformer \cite{vaswani2017attention} and ADAT \cite{shahin2026adat} for translation. ST-GCN++ is a Spatio-Temporal Graph Convolutional Network originally introduced for skeleton-based action recognition. It models human motion as a sequence of spatiotemporal graphs, where joints serve as nodes and physical connections as edges. The model stacks ten spatial graph convolution layers with multi-scale temporal convolutions using parallel dilated kernels. This design captures global motion patterns and fine-grained dynamics with low computational cost. Although it was not designed for sign language, its strong generalization on large-scale skeleton-action benchmarks makes it a suitable backbone for SLMT.

The Transformer serves as a strong baseline as it is the most widely used and accurate model for SLMT \cite{shahin2024rule}. Its encoder–decoder structure separates visual encoding from linguistic generation, and its multi-head attention layers provide full pairwise context modeling. However, the quadratic attention cost makes it computationally demanding for long video sequences. In addition, it does not efficiently capture short, fine-grained temporal dependencies in signs.

ADAT is an adaptive time-series-aware Transformer-based model. It consists of a dual-branch encoder and a decoder. The encoder separates local and global temporal processing to capture fine-grained, short-range motion patterns and long-range dependencies at reduced computational cost. An adaptive gating mechanism dynamically balances the contributions of both branches. This design processes rapid and fine-grained motions while preserving context. Consequently, it provides a more efficient, temporally sensitive alternative to the Transformer for SLMT.

\subsubsection{Evaluation Metrics}
To provide a comparable evaluation of translation quality and computational efficiency, we assess translation quality using Bilingual Evaluation Understudy (BLEU) \cite{papineni2002bleu} and Recall-Oriented Understudy for Gisting Evaluation (ROUGE) \cite{see2017get} performance and measure efficiency through training time.

BLEU captures n-gram precision with a length penalty to measure length mismatch and fluency. It is computed using the geometric mean of n-gram precisions with a brevity penalty \((BP)\), as defined in Equations~\ref{eq:bleu}.
\[
BLEU= \left(\prod_{n=1}^{N} p_n^{\,w_n}\right), \quad
\text{BP} =
\begin{cases}
1, & \text{if } c > r, \\[6pt]
e^{\,1-\frac{r}{c}}, & \text{if } c \le r.
\end{cases}
\tag{12}\label{eq:bleu}
\]

 where \(p_n\) is the precision of n-grams, \(w_n\) is the weight of each n-gram size, \(c\) is the length of the candidate translation, and \(r\) is the length of the reference sequence.
 
ROUGE measures recall of overlapping sequences based on the longest common subsequence \((LCS)\) between a generated \((G)\) and a reference translations \((R)\). The ROUGE-L captures precision and recall and is given by Equation~\ref{eq:rouge}.

\[
\text{ROUGE-L} =
\frac{2 \cdot \text{LCS}(G, R)}
{|G| + |R|}
\tag{13}\label{eq:rouge}
\]

For computational efficiency, we compute training time as the total number of hours required to complete model training under the unified experimental setup.

\subsection{Experiments}
To evaluate HATL, we compare it against two fine-tuning baselines: 1) classical fine-tuning, where only the translation model is trained while the backbone remains frozen, and 2) full fine-tuning, where all backbone layers and the translation model are unfrozen and trained from the start. We conduct these comparisons using two translation models: the standard Transformer and ADAT. 

To ensure fair comparison, we use the same backbone across all configurations and evaluate all models under a unified setup for Sign2Text and Sign2Gloss2Text translation tasks on PHOENIX14T, Isharah, and MedASL datasets. We conduct all experiments in PyTorch using 2 NVIDIA RTX A6000 GPUs. 

We perform a structured, multi-stage hyperparameter search, tuning parameter subsets at each stage while keeping the rest fixed. Table~\ref{tab:hparams} summarizes all hyperparameters used to evaluate HATL and the translation models, respectively.

\begin{table}
\caption{Hyperparameters used in all experiments.}
\label{tab:hparams}
\centering
\setlength{\tabcolsep}{3pt}
\renewcommand{\arraystretch}{1.15}
\begin{tabular}{p{0.42\columnwidth} p{0.50\columnwidth}}
\toprule
Hyperparameter 
& Value Used \\
\midrule
\textit{HATL} & \\

Warmup epochs & 
$2$ \\

Warmup scheduler & 
$\max(200; 
0.02 \times \text{training steps})$ \\

Patience for unfreezing the first layer & 
$4$ \\

Moving average window & 
$3$ epochs \\

Unfreeze thresholds & 
CTC: $0.003$; BLEU-4: $0.002$ \\

Unfreeze thresholds decay & 
$\times 0.95$ every $5$ epochs \\

Cooldown after unfreeze & 
$3$ epochs \\

Early stopping & 
$5$ epochs after cooldown\\

Optimizer & 
AdamW, $(0.9,0.98)$, $\epsilon=10^{-8}$ \\

Backbone learning rate & 
$1{\times}10^{-5}$ \\

layer-wise learning rate decay & 
$0.1 \times (1/2^d)$ \\
\hline
\textit{Translation models} & \\

Encoder layers & $3$ \\

Decoder layers & $1$ \\

Hidden size & $512$ \\

Attention heads & $8$ \\

Dropout rate & $0.1$ \\

Optimizer & 
AdamW, $(0.9,0.98)$, $\epsilon=10^{-8}$ \\

Learning rate &
Encoder: $5 \times 10^{-5}$, decoder: $1 \times 10^{-4}$ \\

CTC gloss alignment & Enabled in Sign2Gloss2Text, disabled in Sign2Text ($\omega_{CTC}=0$) \\

CTC blank penalties in Sign2Gloss2Text & Bias: $0.4$; Temp: $0.9$ \\

Beam search & Beam width: $8$ \\

KenLM Language-model scoring for text beam search & 4-gram LM, weight $0.7$ \\
\midrule
\bottomrule
\end{tabular}
\end{table}

\subsection{Experimental Results Analysis}

This section presents a comprehensive evaluation of HATL across ADAT and Transformer models on Sign2Gloss2Text and Sign2Text translation tasks, comparing it against classical and full static fine-tuning in terms of translation quality and computational efficiency. 

\subsubsection{Translation Quality}
Tables ~\ref{tab:phoenix14t_results}-\ref{tab:medasl_results} report the translation quality results on PHOENIX14T, Isharah, and MedASL datasets, respectively, for Sign2Gloss2Text and Sign2Text. State-of-the-art results are reported for comparison. 

\begin{table*}
\caption{Performance Comparison of Sign Language Translation Models on PHOENIX14T Dataset. Best results are in bold}
\label{tab:phoenix14t_results}
\centering

\begin{adjustbox}{width=\linewidth,totalheight=0.95\textheight,keepaspectratio}

\begin{tabular}{lrrrrrrrrrr}
\hline
\multirow{2}{*}{Model}  &
\multicolumn{5}{c}{Dev} &
\multicolumn{5}{c}{Test} \\
 &
  BLEU-1 &
  BLEU-2 &
  BLEU-3 &
  BLEU-4 &
  ROUGE &
  BLEU-1 &
  BLEU-2 &
  BLEU-3 &
  BLEU-4 &
  ROUGE \\
\hline
\multicolumn{11}{l}{\textit{Sign2Gloss2Text}} \\
NSLT \cite{camgoz2018neural} &
  42.9 &
  30.3 &
  23.0 &
  18.4 &
  44.1 &
  43.3 &
  30.4 &
  22.8 &
  18.1 &
  43.8 \\
Multi-View SLT \cite{li2022sign} &
  - &
  - &
  - &
  - &
  - &
  46.1 &
  32.8 &
  25.0 &
  20.1 &
  - \\
Joint-SLRT \cite{camgoz2020sign} &
  47.3 &
  34.4 &
  27.1 &
  22.4 &
  - &
  46.6 &
  33.7 &
  26.2 &
  21.3 &
  - \\
HST-GNN \cite{kan2022sign} &
  46.1 &
  33.4 &
  27.5 &
  22.6 &
  - &
  45.2 &
  34.7 &
  27.1 &
  22.3 &
  - \\
SimulSLT \cite{yin2021simulslt} &
  47.8 &
  35.3 &
  27.9 &
  22.9 &
  49.2 &
  48.2 &
  35.6 &
  28.0 &
  23.1 &
  49.2 \\
STM-Net \cite{zhou2021spatial} &
  47.6 &
  36.4 &
  29.2 &
  24.1 &
  48.2 &
  47.0 &
  36.1 &
  28.7 &
  23.7 &
  46.7 \\
STMC-Transformer \cite{yin-read-2020-better} &
  46.8 &
  35.0 &
  27.8 &
  23.1 &
  47.3 &
  47.5 &
  35.9 &
  28.6 &
  23.8 &
  47.3 \\
VL-Transfer \cite{chen2022simple} &
  50.4 &
  37.5 &
  29.7 &
  24.6 &
  50.2 &
  49.9 &
  37.3 &
  29.7 &
  24.6 &
  49.6 \\
SLTUNET \cite{zhang2023sltunet} &
  - &
  - &
  - &
  25.4 &
  49.6 &
  50.4 &
  39.2 &
  31.4 &
  26.0 &
  50.0 \\
Two Stream-SLT \cite{chen2022two} &
  52.4 &
  39.8 &
  31.9 &
  26.5 &
  52.0 &
  52.1 &
  39.8 &
  32.0 &
  26.7 &
  51.6 \\
Transformer (Classical Fine-tuning) &
 42.7 &
 31.5 &
 24.4 &
 19.8 &
 42.2 &
 41.3 &
 30.3 &
 23.3 &
 18.8 &
 41.7 \\
Transformer (Full Fine-tuning) &
 43.6 &
 32.0 &
 24.7 &
 20 &
 43.5 &
 42.0 &
 31.0 &
 23.8 &
 19.7 &
 43.2 \\
Transformer (HATL) &
 68.0 &
 49.1 &
 36.6 &
 26.8 &
 49.8 &
 67.7 &
 48.2 &
 35.5 &
 25.7 &
 50.5 \\
ADAT (Classical Fine-tuning) &
 42.2 &
 31.1 &
 24.3 &
 19.9 &
 42.4 &
 39.7 &
 29.3 &
 22.8 &
 18.5 &
 40.6 \\
ADAT (Full Fine-tuning) &
 42.8 &
 31.2 &
 24.2 &
 20.4 &
 42.8 &
 41.0 &
 30.6 &
 23.6 &
 20.2 &
 43.5 \\
ADAT (HATL) &
\textbf{ 70.7 }&
\textbf{ 51.0 }&
\textbf{ 38.1 }&
\textbf{ 35.1 }&
\textbf{ 54.5 }&
\textbf{ 68.0 }&
\textbf{ 48.4 }&
\textbf{ 35.6} &
\textbf{ 30.7 }&
\textbf{ 52.8} \\
\hline
\multicolumn{11}{l}{\textit{Sign2Text}} \\
NSLT \cite{camgoz2018neural} &
  31.9 &
  19.1 &
  13.2 &
  9.9 &
  31.8 &
  32.2 &
  19.9 &
  12.8 &
  9.6 &
  31.8 \\
Multi-View SLT \cite{li2022sign} &
  - &
  - &
  - &
  - &
  - &
  34.4 &
  21.0 &
  14.6 &
  11.2 &
  - \\
SimulSLT \cite{yin2021simulslt} &
  - &
  - &
  - &
  - &
  - &
  - &
  - &
  - &
  12.3 &
  - \\
SignNet II \cite{chaudhary2022signnet} &
  - &
  - &
  - &
  - &
  - &
  39.2 &
  24.6 &
  16.9 &
  12.3 &
  - \\
MC-SLT \cite{jin2022mc} &
  - &
  - &
  - &
  - &
  - &
  43.7 &
  - &
  - &
  17.0 &
  43.4 \\
Joint-SLRT \cite{camgoz2020sign} &
  45.5 &
  32.6 &
  25.3 &
  20.7 &
  - &
  45.3 &
  32.3 &
  24.8 &
  20.2 &
  - \\
STMC-Transformer \cite{yin-read-2020-better} &
  50.3 &
  37.6 &
  29.8 &
  24.7 &
  48.7 &
  50.6 &
  38.4 &
  30.6 &
  25.4 &
  48.8 \\
VL-Transfer \cite{chen2022simple} &
  54.0 &
  41.1 &
  33.1 &
  27.6 &
  53.1 &
  54.0 &
  41.8 &
  33.8 &
  28.4 &
  52.7 \\
SLTUNET \cite{zhang2023sltunet} &
  - &
  - &
  - &
  27.9 &
  52.2 &
  52.9 &
  41.8 &
  34.0 &
  28.5 &
  52.1 \\
Two Stream-SLT \cite{chen2022two} &
  54.3 &
  42.0 &
  34.2 &
\textbf{  28.7} &
  54.1 &
  54.9 &
  42.4 &
  34.5 &
  29.0 &
  53.5 \\
Transformer (Classical Fine-tuning) &
  42.7 &
  31.1 &
  24.2 &
  16.2 &
  36.4 &
  41.0 &
  29.9 &
  23.1 &
  16.4 &
  36.8 \\
Transformer (Full Fine-tuning) &
  42.9 &
  31.4 &
  24.4 &
  17.7 &
  37.3 &
  41.2 &
  31.2 &
  23.4 &
  17.5 &
  37.1 \\
Transformer (HATL) &
  45.5 &
  31.7 &
  22.8 &
  19.8 &
  43.9 &
  45.1 &
  31.5 &
  22.5 &
  18.6 &
  42.6 \\
ADAT (Classical Fine-tuning) &
  37.0 &
  28.3 &
  22.7 &
  18.0 &
  42.5 &
  36.0 &
  27.6 &
  21.8 &
  17.0 &
  41.1 \\
ADAT (Full Fine-tuning) &
  38.3 &
  29.2 &
  23.5 &
  19.7 &
  44.1 &
  37.0 &
  28.0 &
  22.4 &
  18.6 &
  42.8 \\
ADAT (HATL) &
\textbf{   60.5} &
\textbf{   50.8 }&
\textbf{   38.0 }&
   28.6 &
\textbf{   54.6} &
\textbf{   59.5} &
\textbf{   48.2} &
\textbf{   35.4} &
\textbf{   29.4 }&
\textbf{   54.2} \\
\hline
\end{tabular}
\end{adjustbox}
\end{table*}

\begin{table*}
\caption{Performance Comparison of Sign Language Translation Models on Isharah Dataset. Best results are in bold}
\label{tab:isharah_results}
\centering
\begin{adjustbox}{width=\linewidth,totalheight=0.95\textheight,keepaspectratio,center}

\begin{tabular}{lrrrrrrrrrr}
\toprule
\multirow{2}{*}{Model}  &
\multicolumn{5}{c}{Dev} &
\multicolumn{5}{c}{Test} \\
 &
  BLEU-1 &
  BLEU-2 &
  BLEU-3 &
  BLEU-4 &
  ROUGE &
  BLEU-1 &
  BLEU-2 &
  BLEU-3 &
  BLEU-4 &
  ROUGE \\
\midrule
\multicolumn{11}{l}{\textit{Sign2Gloss2Text}} \\

VL-Transfer \cite{chen2022simple} & 
71.8 & 
70.8 & 
68.0 & 
66.1 & 
72.3 & 
51.3 & 
49.9 & 
48.3 & 
45.4 & 
52.6 \\
Transformer (Classical Fine-tuning) & 
57.5 & 
47.3 & 
43.2 & 
40.9 & 
55.8 & 
64.1 & 
52.1 & 
45.7 & 
40   & 
45.2 \\
Transformer (Full Fine-tuning)  & 
57.9 & 
47.8 & 
43.4 & 
41.4 & 
60.0 & 
63.7 & 
52.4 & 
46.5 & 
41.5 & 
45.9 \\
Transformer (HATL) & 
80.3 & 
64.3 & 
54.6 & 
45.5 & 
74.8 & 
88.7 & 
68.2 & 
56.3 & 
45.6 & 
53.4 \\
ADAT (Classical Fine-tuning) & 
57.5 & 
47.0 & 
42.3 & 
41.7 & 
56.3 & 
61.4 & 
51.4 & 
44.2 & 
40.9 & 
35.2 \\
ADAT (Full Fine-tuning) & 
57.2 & 
47.4 & 
41.9 & 
42.8 & 
60.1 & 
63.1 & 
52.3 & 
46.1 & 
42.4 & 
40.1 \\
ADAT (HATL)  & 
\textbf{85.1} & 
\textbf{66.2} & 
\textbf{55.5} & 
\textbf{57.2} & 
\textbf{75.3} & 
\textbf{90.4 }& 
\textbf{70.1 }& 
\textbf{57.6} & 
\textbf{52.2 }& 
\textbf{55.5 }\\
\hline
\multicolumn{11}{l}{\textit{Sign2Text}} \\
GFSLT-VLP \cite{zhou2023gloss}&
68.0 &
66.3 &
65.3 &
64.8 &
69.1 & 
47.8 &
45.8 &
44.6 &
43.4 &
49.5 \\
Transformer (Classical Fine-tuning) &
55.3 &
45.6 &
41.3 &
39.1 &
54.1 &
61.0 &
49.1 &
43.1 &
38.3 &
43.2 \\
Transformer (Full Fine-tuning) &
55.5 &
45.5 &
41.4 &
39.5 &
57.4 &
60.5 &
49.5 &
44.1 &
39.1 &
44.4 \\
Transformer (HATL) &
76.2 &
61.0 &
52.1 &
43.0 &
71.9 &
84.1 &
65.2 &
53.1 &
43.3 &
51.1 \\
ADAT (Classical Fine-tuning) &
55.1 &
44.5 &
40.9 &
39.2 &
54.3 &
59.3 &
48.2 &
41.2 &
39.1 &
33.3 \\
ADAT (Full Fine-tuning) &
55.0 &
44.3 &
40.8 &
40.0 &
58.1 &
59.9 &
48.7 &
44.1 &
39.9 &
35.8 \\
ADAT (HATL) &
\textbf{82.0 }&
\textbf{63.2 }&
\textbf{53.1 }&
\textbf{55.2 }&
\textbf{73.1 }&
\textbf{86.3 }&
\textbf{66.9 }&
\textbf{55.3} &
\textbf{49.5 }&
\textbf{53.2 }\\
\midrule
\bottomrule

\end{tabular}

\end{adjustbox}\end{table*}

\begin{table*}
\caption{Performance Comparison of Sign Language Translation Models on MedASL Dataset. Best results are in bold}
\label{tab:medasl_results}
\begin{adjustbox}{width=\linewidth,totalheight=0.95\textheight,keepaspectratio,center}
\begin{tabular}{lrrrrrrrrrr}
\toprule
\multirow{2}{*}{Model}  &
\multicolumn{5}{c}{Dev} &
\multicolumn{5}{c}{Test} \\
 &
  BLEU-1 &
  BLEU-2 &
  BLEU-3 &
  BLEU-4 &
  ROUGE &
  BLEU-1 &
  BLEU-2 &
  BLEU-3 &
  BLEU-4 &
  ROUGE \\
\midrule
\multicolumn{11}{l}{\textit{Sign2Gloss2Text}} \\

Transformer (Classical Fine-tuning)        & 
51.8 & 
41.9 & 
34.8 & 
28.8 & 
51.0 & 
47.9 & 
38.6 & 
32.0 & 
26.3 & 
50.6 \\
Transformer (Full Fine-tuning)             & 
51.6 & 
42.6 & 
36.4 & 
31.2 & 
52.5 & 
50.8 & 
41.6 & 
34.8 & 
29.2 & 
51.8 \\
Transformer (HATL)                         & 
81.5 & 
63.3 & 
50.8 & 
39.1 & 
53.2 & 
79.1 & 
60.9 & 
48.3 & 
36.8 & 
51.8 \\
ADAT (Classical Fine-tuning)               & 
55.2 & 
44.7 & 
37.1 & 
30.5 & 
52.0 & 
50.8 & 
40.5 & 
33.3 & 
27.6 & 
50.9 \\
ADAT (Full Fine-tuning)                    & 
54.2 & 
45.1 & 
38.4 & 
32.7 & 
53.9 & 
52.7 & 
43.5 & 
37.0 & 
31.4 & 
53.2 \\
ADAT (HATL)                                & 
\textbf{77.5 }& 
\textbf{59.6} & 
\textbf{47.6 }& 
\textbf{43.4} & 
\textbf{59.5} & 
\textbf{79.2 }& 
\textbf{59.6 }& 
\textbf{46.0 }& 
\textbf{43.2 }& 
\textbf{59.1 }\\
\hline
\multicolumn{11}{l}{\textit{Sign2Text}} \\
Transformer (Classical Fine-tuning) &
45.0   &
36.5   &
30.7   &
26.0   &
45.6   &
45.4   &
36.1   &
29.9   &
24.8   &
45.9   \\
Transformer (Full Fine-tuning) &
52.4   &
42.0   &
34.7   &
26.9   &
45.8   &
51.3   &
40.5   &
33.1   &
26.1   &
46.2   \\
Transformer (HATL) &
72.6   &
54.2   &
41.8   &
28.3   &
46.1   &
73.3   &
55.1   &
42.8   &
29.1   &
47.2   \\
ADAT (Classical Fine-tuning) &
53.0   &
35.3   &
23.7   &
15.6   &
26.7   &
51.6   &
34.5   &
23.4   &
15.6   &
26.0   \\
ADAT (Full Fine-tuning) &
46.8   &
38.2   &
32.2   &
27.3   &
47.1   &
47.9   &
39.2   &
33.2   &
28.3   &
47.8   \\
ADAT (HATL) &
\textbf{50.4} &
\textbf{41.0} &
\textbf{34.4}  &
\textbf{29.2}  &
\textbf{50.0}  &
\textbf{52.7}  &
\textbf{43.1}   &
\textbf{36.3} &
\textbf{30.5}  &
\textbf{51.6}  \\
\bottomrule
\end{tabular}
\end{adjustbox}
\end{table*}

\textit{1. Sign2Gloss2Text}

Across all datasets, static fine-tuning approaches often converge to similar performance, while HATL consistently outperforms both baselines, demonstrating that progressive hierarchical adaptation is more stable than static approaches.

On PHOENIX14T, static fine-tuning results in comparable performance for the Transformer and ADAT, whereas HATL consistently improves both models. The Transformer shows improvements across all metrics, indicating enhanced gloss-text alignment and sentence-level structure. ADAT benefits more significantly due to its temporally adaptive architecture. It outperforms both fine-tuning baselines with +12.2 and +10.5 BLEU-4, respectively. It also surpasses prior state-of-the-art systems by at least +3.0 BLEU-4, demonstrating the combined effectiveness of ADAT and HATL.

On Isharah and MedASL, HATL achieves significant improvements across all metrics, with ADAT showing the largest gains. In particular, In Isharah, ADAT results in +11.3 and +9.8 BLEU-4 over classical and full fine-tuning, respectively. In MedASL, ADAT outperforms classical and fine-tuning with +15.6 and +11.8 BLEU-4, respectively. The enhancements observed in ROUGE indicate better preservation of domain-specific terminology, which is critical for healthcare SLMT applications.

In summary, HATL consistently outperforms both static fine-tuning approaches across all datasets. Its benefits are most notable in complex and domain-specific environments such as PHOENIX14T, while remaining effective in low-variability settings such as Isharah. The gains are particularly evident in ADAT, as its architecture is more sensitive to temporal structure. By progressively unfreezing layers, HATL preserves pretrained temporal structure while enabling performance-aware hierarchical adaptation, leading to stable improvements across BLEU and ROUGE metrics. Overall, HATL achieves state-of-the-art results on PHOENIX14T and Isharah datasets, with substantial improvements on MedASL, demonstrating robustness across signers and languages.

\textit{2. Sign2Text}

Sign2Text follows the same performance trends as Sign2Gloss2Text. On PHOENIX14T, the Transformer shows consistent improvements across n-grams, while ADAT reveals the full impact of HATL, due to its time-series-aware encoder. HATL enables ADAT to surpass prior Sign2Text state-of-the-art models, including Two-Stream SLT \cite{chen2022two} (+0.4 BLEU-4) and SLTUNET \cite{zhang2023sltunet}(+0.9 BLEU-4).

On Isharah, HATL results in significant improvements in both translation models. ADAT achieves the highest BLEU and ROUGE scores, outperforming the state-of-the-art by +6.0 BLEU-4 and +3.7 ROUGE.

On MedASL, similar to the other results, ADAT benefits the most from HATL, resulting in +14.9 BLEU-4 over classical and +2.2 over full fine-tuning. The significant gains in ROUGE indicates improved content preservation, which is critical in medical translation.

In summary, HATL consistently outperforms static fine-tuning. Its effectiveness is more pronounced for ADAT than the Transformer, leading to more accurate translations than state-of-the-art approaches. PHOENIX14T and MedASL highlight HALT's robustness under linguistic diversity and domain specificity, while Isharah results in higher scores due to its structure. In particular, Isharah includes 15 signers, which is large for keypoint-based models. Keypoint representations reduce visual identity bias, making models lighter and less signer-dependent than RGB-based approaches \cite{shahin2025towards}.

\subsubsection{Computational Efficiency}
\begin{figure}
\centerline{\includegraphics[width=4in]{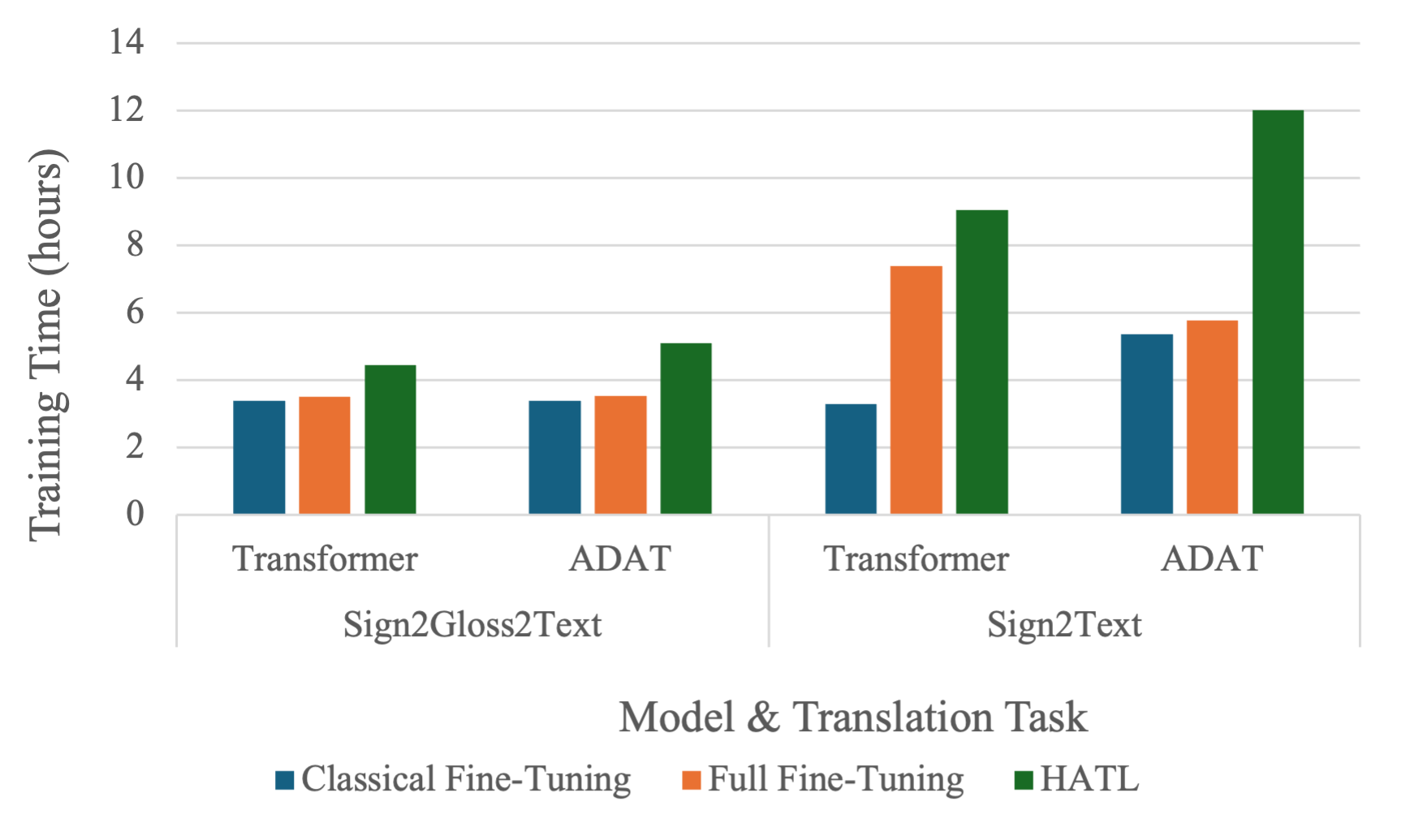}}
\caption{Training time for fine-tuning approaches across Transformer and ADAT models on PHOENIX14T dataset.}
\label{fig:phoenix_training}
\end{figure}

\begin{figure}
\centerline{\includegraphics[width=4in]{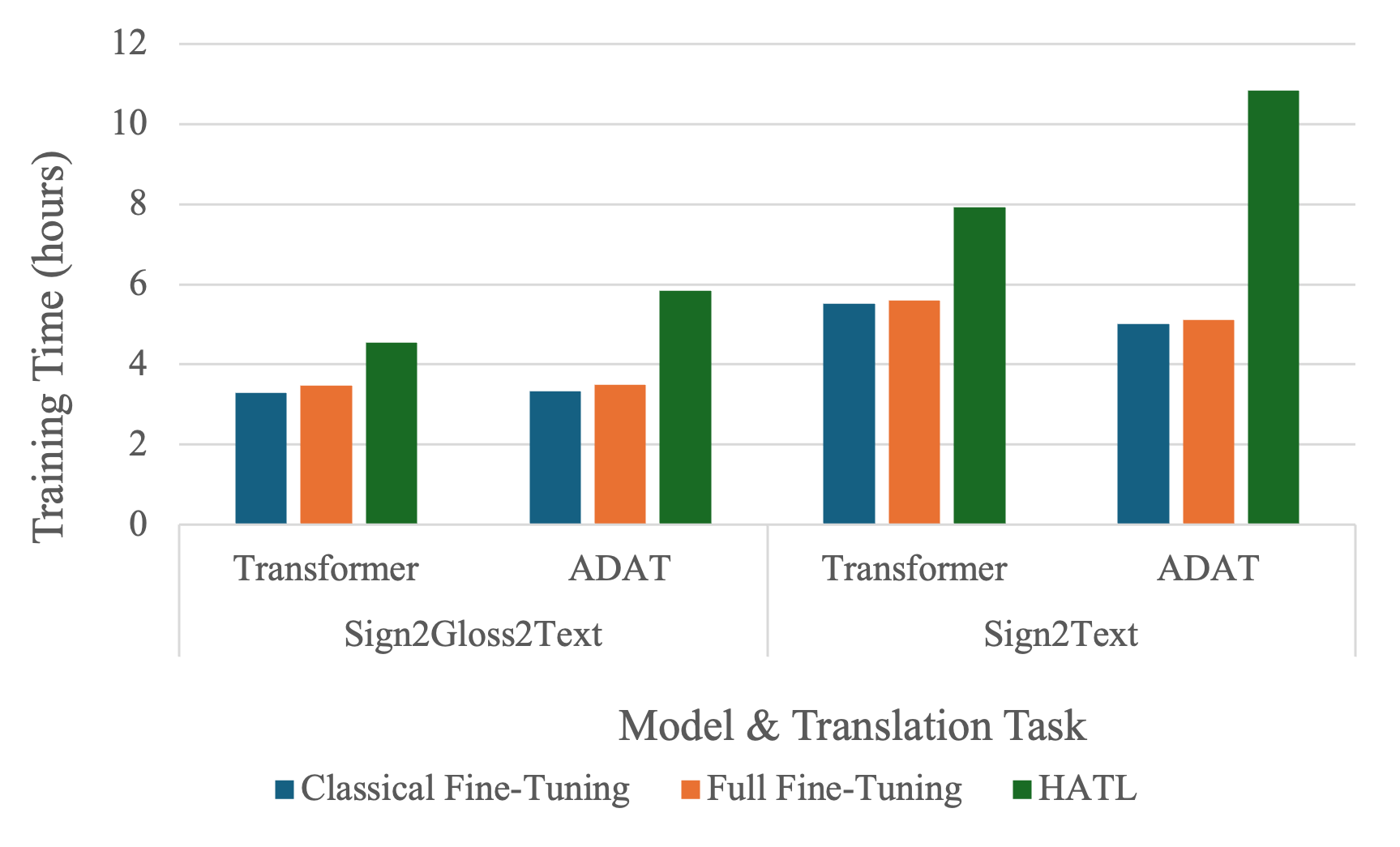}}
\caption{Training time for fine-tuning approaches across Transformer and ADAT models on Isharah dataset.}
\label{fig:isharah_training}
\end{figure}

\begin{figure}
\centerline{\includegraphics[width=4in]{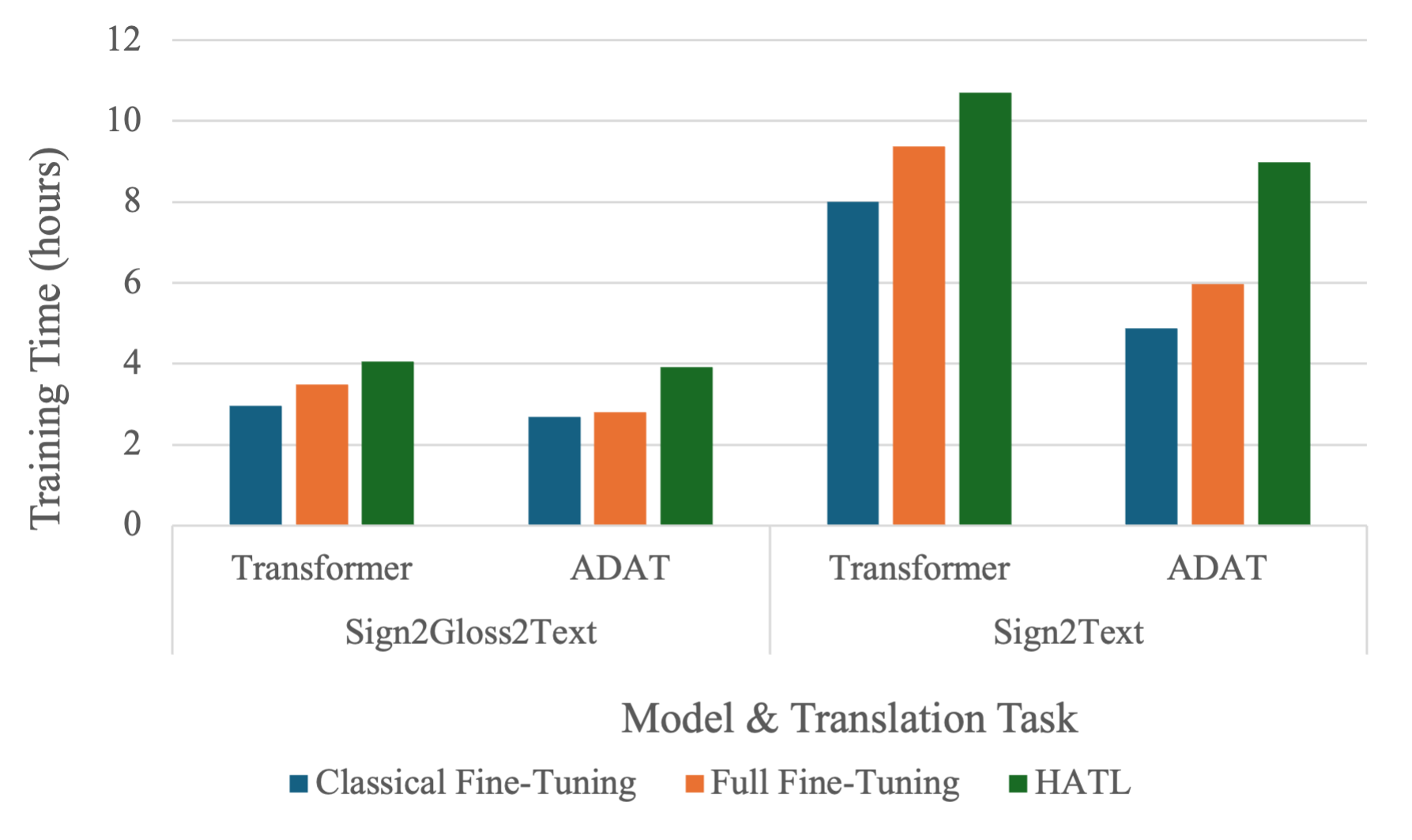}}
\caption{Training time for fine-tuning approaches across Transformer and ADAT models on MedASL dataset.}
\label{fig:medasl_training}
\end{figure}

Figures~\ref{fig:phoenix_training}–\ref{fig:medasl_training} compare the training time in hours across fine-tuning approaches for PHOENIX14T, Isharah, and MedASL datasets, covering all models and translation tasks. Classical fine-tuning is consistently the most efficient, as the backbone remains frozen and only the translation model is updated. Full fine-tuning has higher computational cost, as all backbone layers are updated from the start, resulting in a larger number of trainable parameters. HATL introduces additional overhead due to its hierarchical progressive activation, where each newly unfrozen layer expands the optimization space and increases training duration. These trends are consistent across datasets, translation models, and tasks. 

\begin{figure}
\centerline{\includegraphics[width=4in]{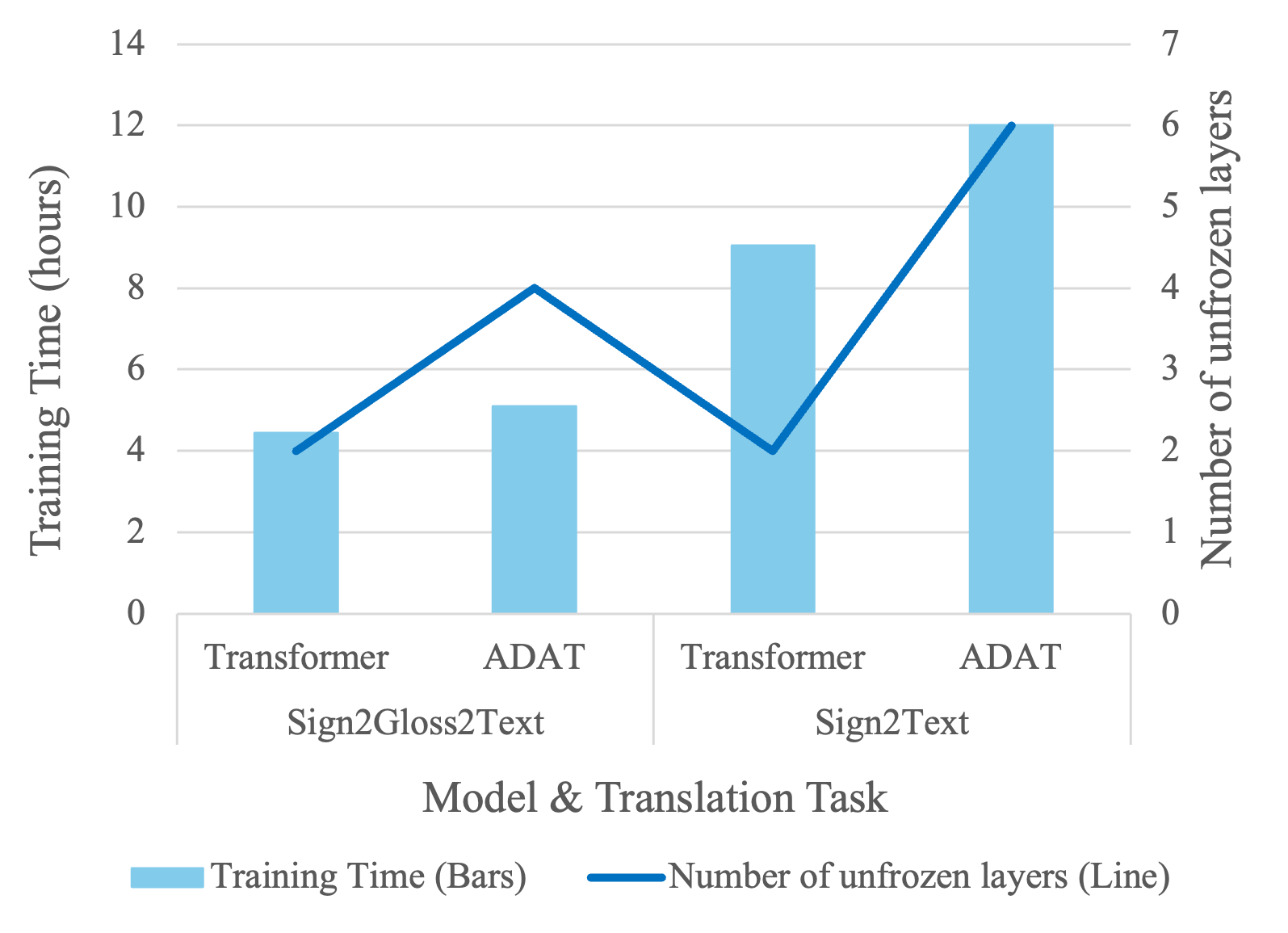}}
\caption{Training time versus number of unfrozen layers using HATL for Transformer and ADAT on PHOENIX14T dataset.}
\label{fig:phoenix_2}
\end{figure}

\begin{figure}

\centerline{\includegraphics[width=4in]{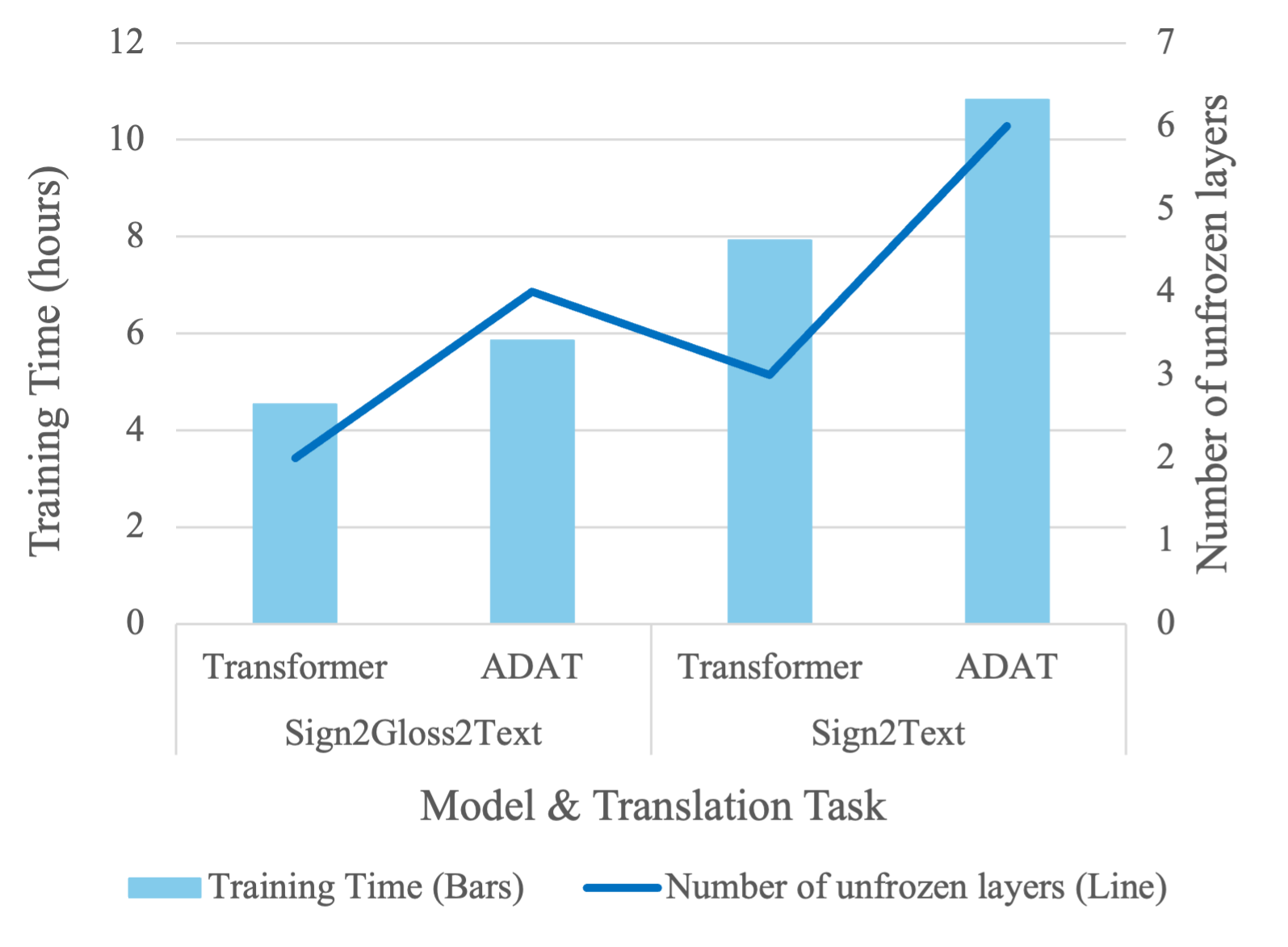}}
\caption{Training time versus number of unfrozen layers using HATL for Transformer and ADAT on Isharah dataset.}
\label{fig:isharah_2}
\end{figure}

\begin{figure}

\centerline{\includegraphics[width=4in]{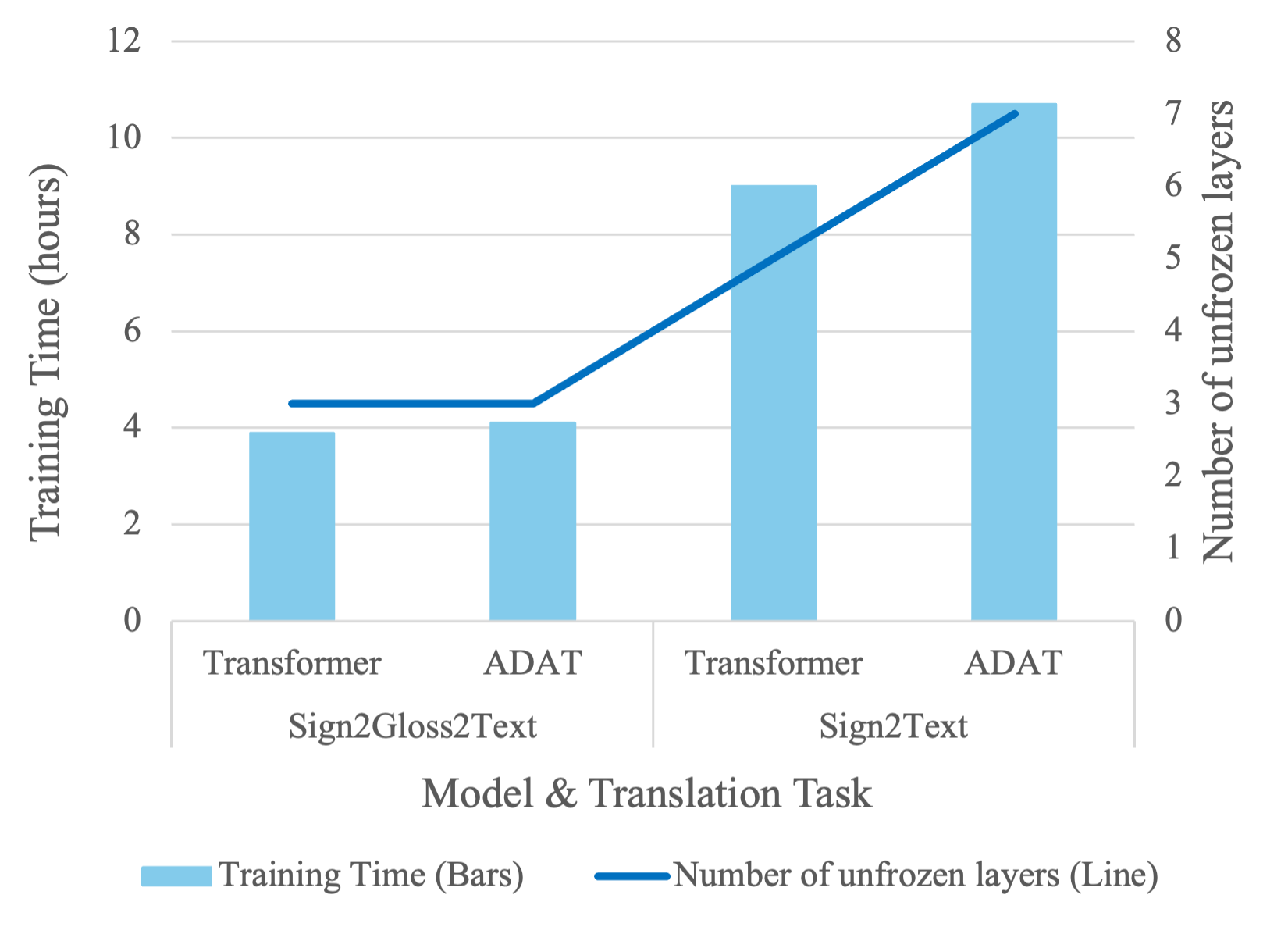}}
\caption{Training time versus number of unfrozen layers using HATL for Transformer and ADAT on MedASL dataset.}
\label{fig:medasl_2}
\end{figure}

\begin{figure}
\centerline{\includegraphics[width=4in]{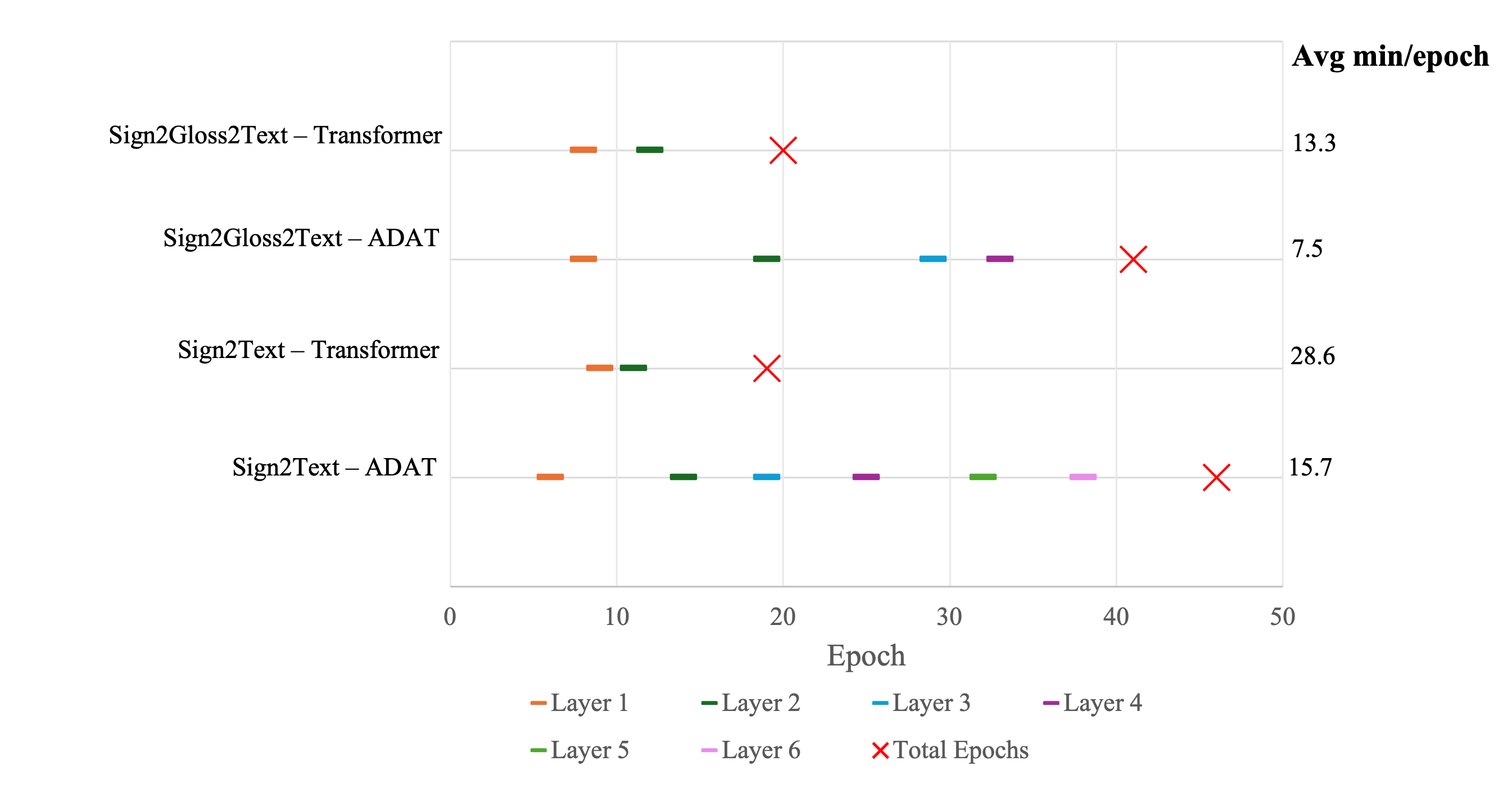}}
\caption{Hierarchical
 unfreezing timelines using HATL for Transformer and ADAT on PHOENIX14T dataset.}
\label{fig:phoenix_3}
\end{figure}

\begin{figure}
\centerline{\includegraphics[width=4in]{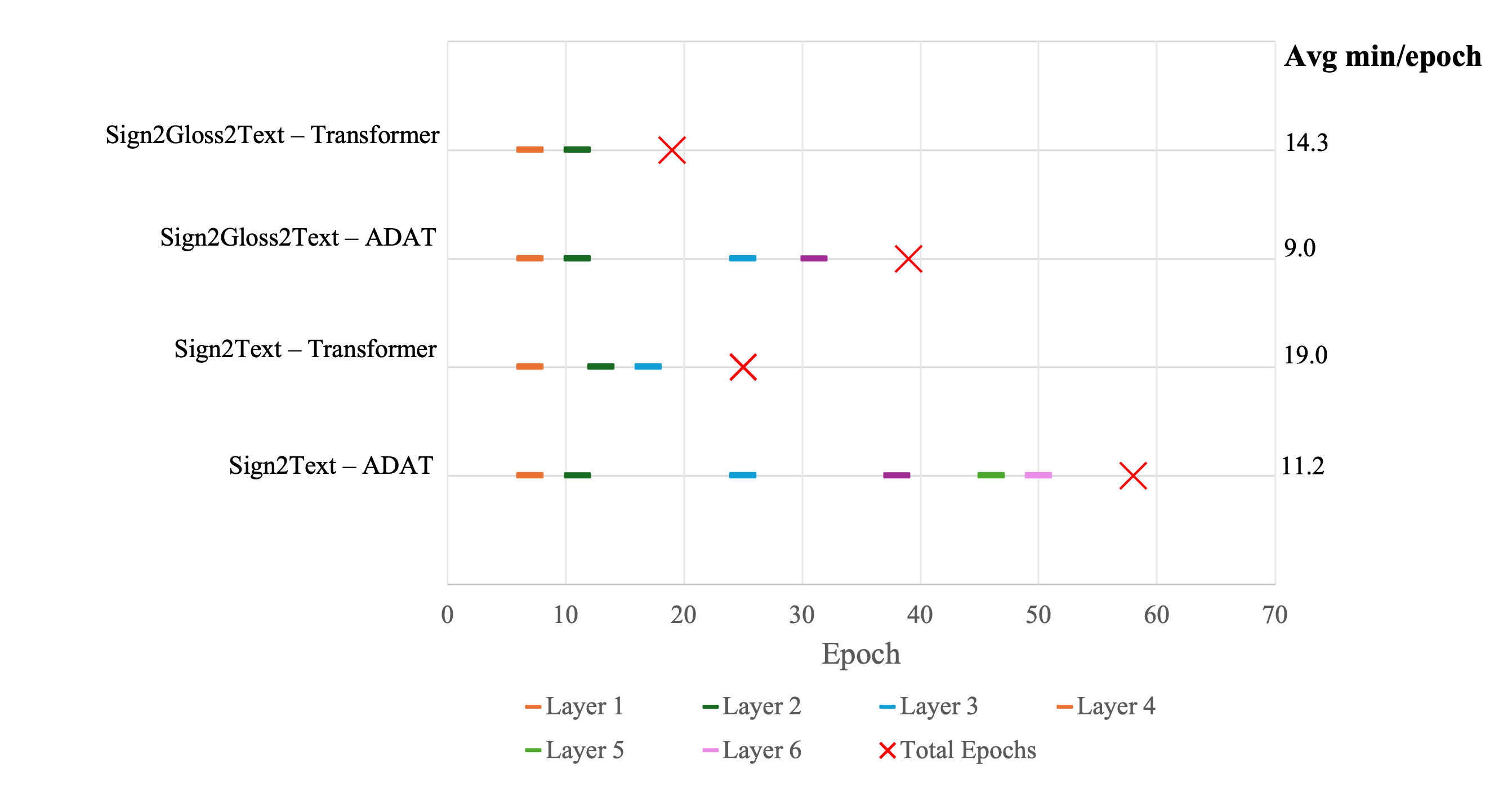}}
\caption{Hierarchical unfreezing timelines using HATL for Transformer and ADAT on Isharah dataset.}
\label{fig:isharah_3}
\end{figure}

\begin{figure}
\centerline{\includegraphics[width=4in]{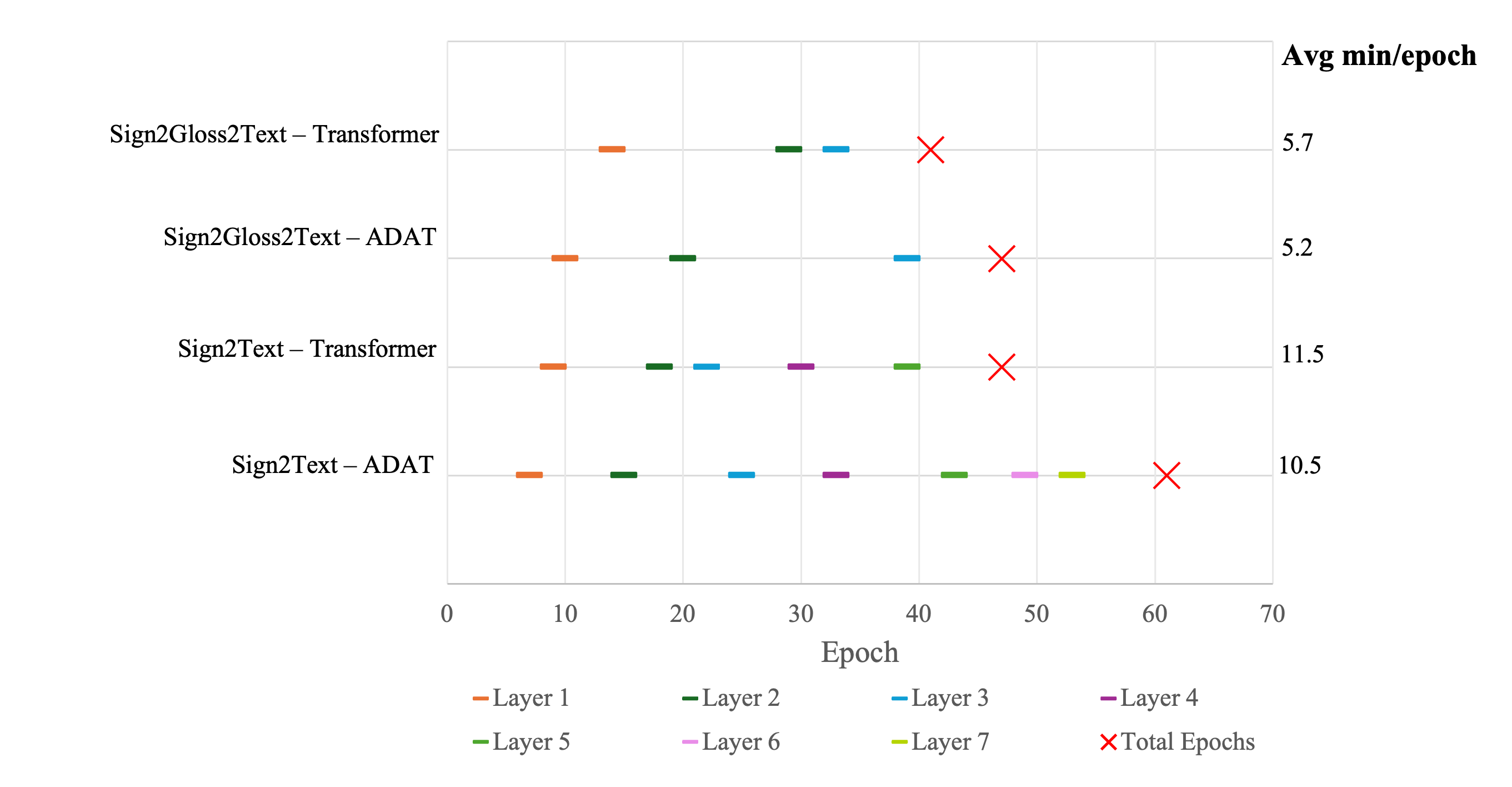}}
\caption{Hierarchical unfreezing timelines using HATL for Transformer and ADAT on MedASL dataset.}
\label{fig:medasl_3}
\end{figure}

Figures~\ref{fig:phoenix_2}-\ref{fig:medasl_3} illustrate the training behavior of Transformer and ADAT models using HATL across all datasets and translation tasks. Across all settings, ADAT consistently unfreezes more pretrained layers than the Transformer. As a result, training extends to later epochs, leading to higher total training time. Nevertheless, ADAT consistently maintains a lower average time per epoch than the Transformer, indicating that the increased training time is due to the extended training rather than reduced per-epoch efficiency.

In summary, computational cost is driven by the extent and duration of backbone adaptation. Classical fine-tuning is the most efficient due to the frozen backbone, while full fine-tuning is more expensive due to simultaneous training of all pretrained layers. HATL increases training time through progressive hierarchical unfreezing. Within this framework, ADAT has higher total training time than the Transformer while maintaining lower average time per epoch.

Overall, the improved translation quality achieved by HATL is linked to its dynamic hierarchical unfreezing, which requires more epochs, increasing total training time. This additional cost enables deeper specialization, resulting in state-of-the-art translation quality.

\section{Conclusion and Future Work}
This paper proposes a performance-aware Hierarchical Adaptive Transfer Learning (HATL) framework for Sign Language Machine Translation (SLMT). HATL progressively increases trainable capacity based on performance, preserving pretrained representations while adapting deeper layers to sign language. Experiments on PHOENIX14T, Isharah, and MedASL datasets across Sign2Text and Sign2Gloss2Text tasks show that HATL consistently surpasses static fine-tuning approaches. Combining HATL with the Adaptive Transformer outperforms transfer learning baselines in the Transformer, highlighting the effectiveness of performance-aware transfer learning for SLMT. Future work should extend HATL to other domains than SLMT and evaluate it using different pretrained models than ST-GCN.

\paragraph{Funding}This research was funded by the Emirates Center for Mobility Research, United Arab Emirates University.

\paragraph{Data Availability} The dataset and the implementation code are publicly available at
\url{https://github.com/INDUCE-Lab/}.
\bibliographystyle{IEEEtran}
\bibliography{references}

@article{hall2019deaf,
  title={Deaf children need language, not (just) speech},
  author={Hall, Matthew L and Hall, Wyatte C and Caselli, Naomi K},
  journal={First language},
  volume={39},
  number={4},
  pages={367--395},
  year={2019},
  publisher={Sage Publications Sage UK: London, England}
}

@inproceedings{chen2022simple,
  title={A simple multi-modality transfer learning baseline for sign language translation},
  author={Chen, Yutong and Wei, Fangyun and Sun, Xiao and Wu, Zhirong and Lin, Stephen},
  booktitle={Proceedings of the IEEE/CVF Conference on Computer Vision and Pattern Recognition},
  pages={5120--5130},
  year={2022}
}

@inproceedings{fu2024signer,
      title = "Signer Diversity-driven Data Augmentation for Signer-Independent Sign Language Translation",
    author = "Fu, Honghao  and
      Zhang, Liang  and
      Fu, Biao  and
      Zhao, Rui  and
      Su, Jinsong  and
      Shi, Xiaodong  and
      Chen, Yidong",
    editor = "Duh, Kevin  and
      Gomez, Helena  and
      Bethard, Steven",
    booktitle = "Findings of the Association for Computational Linguistics: NAACL 2024",
    month = jun,
    year = "2024",
    address = "Mexico City, Mexico",
    publisher = "Association for Computational Linguistics",
    url = "https://aclanthology.org/2024.findings-naacl.140/",
    doi = "10.18653/v1/2024.findings-naacl.140",
    pages = "2182--2193" 
}

@inproceedings{shahin2024glot,
  title={GLoT: A Novel Gated-Logarithmic Transformer for Efficient Sign Language Translation},
  author={Shahin, Nada and Ismail, Leila},
  booktitle={2024 IEEE Future Networks World Forum (FNWF)},
  pages={885--890},
  year={2024},
  organization={IEEE}
}

@inproceedings{holmes2023scarcity,
  title={From scarcity to understanding: Transfer learning for the extremely low resource irish sign language},
  author={Holmes, Ruth and Rushe, Ellen and De Coster, Mathieu and Bonnaerens, Maxim and Satoh, Shin'ichi and Sugimoto, Akihiro and Ventresque, Anthony},
  booktitle={Proceedings of the IEEE/CVF International Conference on Computer Vision},
  pages={2008--2017},
  year={2023}
}

@inproceedings{camgoz2018neural,
  title={Neural sign language translation},
  author={Camgoz, Necati Cihan and Hadfield, Simon and Koller, Oscar and Ney, Hermann and Bowden, Richard},
  booktitle={Proceedings of the IEEE/CVF Conference on Computer Vision and Pattern Recognition},
  pages={7784--7793},
  year={2018}
}

@article{hosna2022transfer,
  title={Transfer learning: a friendly introduction},
  author={Hosna, Asmaul and Merry, Ethel and Gyalmo, Jigmey and Alom, Zulfikar and Aung, Zeyar and Azim, Mohammad Abdul},
  journal={Journal of Big Data},
  volume={9},
  number={1},
  pages={102},
  year={2022},
  publisher={Springer}
}

@article{pan2009survey,
  title={A survey on transfer learning},
  author={Pan, Sinno Jialin and Yang, Qiang},
journal={IEEE Transactions on Knowledge and Data Engineering},
  volume={22},
  number={10},
  pages={1345--1359},
  year={2009},
  publisher={IEEE}
}

@article{liu2022yolov5,
  title={YOLOv5-Tassel: Detecting tassels in RGB UAV imagery with improved YOLOv5 based on transfer learning},
  author={Liu, Wei and Quijano, Karoll and Crawford, Melba M},
  journal={IEEE Journal of Selected Topics in Applied Earth Observations and Remote Sensing},
  volume={15},
  pages={8085--8094},
  year={2022},
  publisher={IEEE}
}

@inproceedings{xiao2022transfer,
  title={Transfer learning from synthetic to real lidar point cloud for semantic segmentation},
  author={Xiao, Aoran and Huang, Jiaxing and Guan, Dayan and Zhan, Fangneng and Lu, Shijian},
  booktitle={Proceedings of the AAAI Conference on Artificial Intelligence},
  volume={36},
  number={3},
  pages={2795--2803},
  year={2022}
}

@inproceedings{xue2021transfer,
  title={Transfer: Learning relation-aware facial expression representations with transformers},
  author={Xue, Fanglei and Wang, Qiangchang and Guo, Guodong},
  booktitle={Proceedings of the IEEE/CVF International Conference on Computer Vision},
  pages={3601--3610},
  year={2021}
}

@article{soleimani2021cross,
  title={Cross-subject transfer learning in human activity recognition systems using generative adversarial networks},
  author={Soleimani, Elnaz and Nazerfard, Ehsan},
  journal={Neurocomputing},
  volume={426},
  pages={26--34},
  year={2021},
  publisher={Elsevier}
}

@inproceedings{sung2022vl,
  title={Vl-adapter: Parameter-efficient transfer learning for vision-and-language tasks},
  author={Sung, Yi-Lin and Cho, Jaemin and Bansal, Mohit},
  booktitle={Proceedings of the IEEE/CVF Conference on Computer Vision and Pattern Recognition},
  pages={5227--5237},
  year={2022}
}

@article{al2025novel,
  title={Novel transfer learning based acoustic feature engineering for scene fake audio detection},
  author={Al-Shamayleh, Ahmad Sami and Riasat, Hafsa and Alluhaidan, Ala Saleh and Raza, Ali and El-Rahman, Sahar A and AbdElminaam, Diaa Salama},
  journal={Scientific Reports},
  volume={15},
  number={1},
  pages={8066},
  year={2025},
  publisher={Nature Publishing Group UK London}
}

@inproceedings{cooper2020zero,
  title={Zero-shot multi-speaker text-to-speech with state-of-the-art neural speaker embeddings},
  author={Cooper, Erica and Lai, Cheng-I and Yasuda, Yusuke and Fang, Fuming and Wang, Xin and Chen, Nanxin and Yamagishi, Junichi},
  booktitle={ICASSP 2020-2020 IEEE International Conference on Acoustics, Speech and Signal Processing (ICASSP)},
  pages={6184--6188},
  year={2020},
  organization={IEEE}
}

@inproceedings{camgoz2020sign,
  title={Sign language transformers: Joint end-to-end sign language recognition and translation},
  author={Camgoz, Necati Cihan and Koller, Oscar and Hadfield, Simon and Bowden, Richard},
  booktitle={Proceedings of the IEEE/CVF Conference on Computer Vision and Pattern Recognition},
  pages={10023--10033},
  year={2020}
}

@article{vaswani2017attention,
  title={Attention is all you need},
  author={Vaswani, Ashish and Shazeer, Noam and Parmar, Niki and Uszkoreit, Jakob and Jones, Llion and Gomez, Aidan N and Kaiser, {\L}ukasz and Polosukhin, Illia},
  journal={Advances in Neural Information Processing Systems},
  volume={30},
  year={2017}
}

@article{said2025adaptive,
  title={Adaptive Transformer-Based Deep Learning Framework for Continuous Sign Language Recognition and Translation},
  author={Said, Yahia and Boubaker, Sahbi and Altowaijri, Saleh M and Alsheikhy, Ahmed A and Atri, Mohamed},
  journal={Mathematics},
  volume={13},
  number={6},
  pages={909},
  year={2025},
  publisher={MDPI}
}

@article{li2022sign,
  title={Sign language recognition and translation network based on multi-view data},
  author={Li, Ronghui and Meng, Lu},
  journal={Applied Intelligence},
  volume={52},
  number={13},
  pages={14624--14638},
  year={2022},
  publisher={Springer}
}

@inproceedings{kan2022sign,
  title={Sign language translation with hierarchical spatio-temporal graph neural network},
  author={Kan, Jichao and Hu, Kun and Hagenbuchner, Markus and Tsoi, Ah Chung and Bennamoun, Mohammed and Wang, Zhiyong},
  booktitle={Proceedings of the IEEE/CVF winter Conference on Applications of Computer Vision},
  pages={3367--3376},
  year={2022}
}

@inproceedings{yin2021simulslt,
  title={Simulslt: End-to-end simultaneous sign language translation},
  author={Yin, Aoxiong and Zhao, Zhou and Liu, Jinglin and Jin, Weike and Zhang, Meng and Zeng, Xingshan and He, Xiaofei},
  booktitle={Proceedings of the 29th ACM International Conference on Multimedia},
  pages={4118--4127},
  year={2021}
}

@article{zhou2021spatial,
  title={Spatial-temporal multi-cue network for sign language recognition and translation},
  author={Zhou, Hao and Zhou, Wengang and Zhou, Yun and Li, Houqiang},
  journal={IEEE Transactions on Multimedia},
  volume={24},
  pages={768--779},
  year={2021},
  publisher={IEEE}
}

@inproceedings{yin-read-2020-better,
    title = "Better Sign Language Translation with {STMC}-Transformer",
    author = "Yin, Kayo  and
      Read, Jesse",
    editor = "Scott, Donia  and
      Bel, Nuria  and
      Zong, Chengqing",
    booktitle = "Proceedings of the 28th International Conference on Computational Linguistics",
    month = dec,
    year = "2020",
    address = "Barcelona, Spain (Online)",
    publisher = "International Committee on Computational Linguistics",
    url = "https://aclanthology.org/2020.coling-main.525/",
    doi = "10.18653/v1/2020.coling-main.525",
    pages = "5975--5989"
}

@inproceedings{
zhang2023sltunet,
title={{SLTUNET}: A Simple Unified Model for Sign Language Translation},
author={Biao Zhang and Mathias M{\"u}ller and Rico Sennrich},
booktitle={The Eleventh International Conference on Learning Representations},
year={2023}
}

@article{chen2022two,
  title={Two-stream network for sign language recognition and translation},
  author={Chen, Yutong and Zuo, Ronglai and Wei, Fangyun and Wu, Yu and Liu, Shujie and Mak, Brian},
  journal={Advances in Neural Information Processing Systems},
  volume={35},
  pages={17043--17056},
  year={2022}
}

@article{shahin2024rule,
  title={From rule-based models to deep learning transformers architectures for natural language processing and sign language translation systems: survey, taxonomy and performance evaluation},
  author={Shahin, Nada and Ismail, Leila},
  journal={Artificial Intelligence Review},
  volume={57},
  number={10},
  pages={271},
  year={2024},
  publisher={Springer}
}

@article{ismail2025visioslr,
  title={VisioSLR: A Vision Data-Driven Framework for Sign Language Video Recognition and Performance Evaluation on Fine-Tuned YOLO Models},
  author={Ismail, Leila and Shahin, Nada and Tesfaye, Henok and Hennebelle, Alain},
  journal={Procedia Computer Science},
  volume={257},
  pages={85--92},
  year={2025},
  publisher={Elsevier}
}

@article{chaudhary2022signnet,
  title={Signnet ii: A transformer-based two-way sign language translation model},
  author={Chaudhary, Lipisha and Ananthanarayana, Tejaswini and Hoq, Enjamamul and Nwogu, Ifeoma},
  journal={IEEE Transactions on Pattern Analysis and Machine Intelligence},
  volume={45},
  number={11},
  pages={12896--12907},
  year={2022},
  publisher={IEEE}
}

@inproceedings{jin2022mc,
  title={Mc-slt: Towards low-resource signer-adaptive sign language translation},
  author={Jin, Tao and Zhao, Zhou and Zhang, Meng and Zeng, Xingshan},
  booktitle={Proceedings of the 30th ACM International Conference on Multimedia},
  pages={4939--4947},
  year={2022}
}

@inproceedings{guo2018hierarchical,
  title={Hierarchical LSTM for sign language translation},
  author={Guo, Dan and Zhou, Wengang and Li, Houqiang and Wang, Meng},
  booktitle={Proceedings of the AAAI conference on Artificial Intelligence},
  volume={32},
  number={1},
  year={2018}
}

@article{guo2019hierarchical,
  title={Hierarchical recurrent deep fusion using adaptive clip summarization for sign language translation},
  author={Guo, Dan and Zhou, Wengang and Li, Anyang and Li, Houqiang and Wang, Meng},
  journal={IEEE Transactions on Image Processing},
  volume={29},
  pages={1575--1590},
  year={2019},
  publisher={IEEE}
}

@article{alyami2025isharah,
  title={Isharah: A Large-Scale Multi-Scene Dataset for Continuous Sign Language Recognition},
  author={Alyami, Sarah and Luqman, Hamzah and Al-Azani, Sadam and Alowaifeer, Maad and Alharbi, Yazeed and Alonaizan, Yaser},
  journal={arXiv preprint arXiv:2506.03615},
  year={2025}
}

@article{shahin2026adat,
    author = {Shahin, Nada and Ismail, Leila},
    title = {ADAT: Time-Series-Aware Adaptive Transformer Architecture for Sign Language Translation},
    journal = {Scientific Reports},
    year = {2026}
}

@inproceedings{howard2018universal,
  title={Universal Language Model Fine-tuning for Text Classification},
  author={Howard, Jeremy and Ruder, Sebastian},
  booktitle={Proceedings of the 56th Annual Meeting of the Association for Computational Linguistics (Volume 1: Long Papers)},
  pages={328--339},
  year={2018}
}

@article{shahin2025towards,
  title={Towards Trustworthy Sign Language Translation System: A Privacy-Preserving Edge--Cloud--Blockchain Approach},
  author={Shahin, Nada and Ismail, Leila},
  journal={Mathematics},
  volume={13},
  number={23},
  pages={3759},
  year={2025},
  publisher={MDPI}
}

@misc{google2024mediapipe,
  title        = {MediaPipe Solutions Guide},
  author       = {{Google AI}},
  year         = {2024},
  howpublished = {\url{https://ai.google.dev/edge/mediapipe/solutions/guide}},
  note         = {Accessed: 2024-11-20}
}

@inproceedings{papineni2002bleu,
  title={Bleu: a method for automatic evaluation of machine translation},
  author={Papineni, Kishore and Roukos, Salim and Ward, Todd and Zhu, Wei-Jing},
  booktitle={Proceedings of the 40th annual meeting of the Association for Computational Linguistics},
  pages={311--318},
  year={2002}
}

@inproceedings{see2017get,
    title = "Get To The Point: Summarization with Pointer-Generator Networks",
    author = "See, Abigail  and
      Liu, Peter J.  and
      Manning, Christopher D.",
    editor = "Barzilay, Regina  and
      Kan, Min-Yen",
    booktitle = "Proceedings of the 55th Annual Meeting of the Association for Computational Linguistics (Volume 1: Long Papers)",
    month = jul,
    year = "2017",
    address = "Vancouver, Canada",
    publisher = "Association for Computational Linguistics",
    url = "https://aclanthology.org/P17-1099/",
    doi = "10.18653/v1/P17-1099",
    pages = "1073--1083"

}

@inproceedings{duan2022pyskl,
  title={Pyskl: Towards good practices for skeleton action recognition},
  author={Duan, Haodong and Wang, Jiaqi and Chen, Kai and Lin, Dahua},
  booktitle={Proceedings of the 30th ACM international Conference on Multimedia},
  pages={7351--7354},
  year={2022}
}

@inproceedings{zhou2023gloss,
  title={Gloss-free sign language translation: Improving from visual-language pretraining},
  author={Zhou, Benjia and Chen, Zhigang and Clap{\'e}s, Albert and Wan, Jun and Liang, Yanyan and Escalera, Sergio and Lei, Zhen and Zhang, Du},
  booktitle={Proceedings of the IEEE/CVF International Conference on Computer Vision},
  pages={20871--20881},
  year={2023}
}

\end{document}